%% file: acl_latex.tex
\pdfoutput=1

\documentclass[11pt]{article}

\usepackage[]{EMNLP2022}

\usepackage{times}
\usepackage{latexsym}
\usepackage{times}
\usepackage{latexsym}
\usepackage{graphicx}
\usepackage{adjustbox}
\usepackage{booktabs}
\usepackage{todonotes}
\usepackage{multirow}
\usepackage{tabularx}
\usepackage{caption}
\usepackage{subcaption}
\usepackage{fancyhdr}
\usepackage{lipsum}
\usepackage{times}
\usepackage{latexsym}
\usepackage{tabularx}
\usepackage{caption}
\usepackage{subcaption}
\usepackage{fancyhdr}
\usepackage{lipsum}
\usepackage{graphicx}
\input{custom-packages}

\usepackage{microtype}

\usepackage[T1]{fontenc}

\usepackage[utf8]{inputenc}
\usepackage{microtype}
\usepackage{amssymb}
\usepackage{pifont}
\title{QAScore - An Unsupervised Unreferenced Metric for the Question Generation Evaluation}

   \author{
  \textbf{Tianbo Ji}$^{1}$
  ~~~~\textbf{Chenyang Lyu}$^{2}$~~~~\textbf{Gareth Jones}$^{1}$~~~~Liting Zhou $^{2}$~~~~Yvette Graham $^{3}$\\
  $^{1}$ ADAPT Centre \\
  $^{2}$ School of Computing, Dublin City University, Ireland \\
  $^{3}$ School of Computer Science and Statistics, Trinity College Dublin, Ireland \\ 
  \texttt{\{tianbo.ji,yvette.graham,gareth.jones\}@adaptcentre.com} \\
    \texttt{chenyang.lyu2@mail.dcu.ie}, \texttt{liting.zhou@dcu.ie} 
}

\begin{document}
\maketitle

\input{0-Abstract}

\input{1-Introduction}

\input{2-Related-Work}

\input{3-Proposed-Metric}

\input{4-Proposed-Human-Evaluation}

\input{4-Experiment-Results}

\input{5-Conclusion}
\bibliography{refs}
\bibliographystyle{acl_natbib}

\end{document}

%% file: custom-packages.tex
\usepackage{bigstrut}
\usepackage{multirow}

\usepackage{amssymb}
\usepackage{pifont} 
\usepackage{amsmath} 

\usepackage{tabularx}
\newcolumntype{Y}{>{\centering\arraybackslash}X}

\usepackage{algorithm} 
\usepackage{algpseudocode} 
\usepackage{soul} 

%% file: 0-Abstract.tex
\begin{abstract}
Question Generation (QG) aims to automate the task
of composing questions for a passage with a set of chosen answers found within the passage. In recent years, the introduction of neural generation models has resulted in substantial improvements of automatically generated questions in terms of quality, especially compared to traditional approaches that employ manually crafted heuristics. However, the metrics commonly applied in QG evaluations have been criticized for their low agreement with human judgement. We therefore propose a new reference-free evaluation metric that has the potential to provide a better mechanism for evaluating QG systems, called QAScore. Instead of fine-tuning a language model to maximize its correlation with human judgements, QAScore evaluates a question by computing the cross entropy according to the probability that the language model can correctly generate the masked words in the answer to that question. Furthermore, we conduct a new crowd-sourcing human evaluation experiment for the QG evaluation to investigate how QAScore and other metrics can correlate with human judgements. Experiments show that QAScore obtains a stronger correlation with the results of our proposed human evaluation method compared to existing traditional word-overlap-based metrics such as BLEU and ROUGE, as well as the existing pretrained-model-based metric BERTScore.
\end{abstract}

%% file: 1-Introduction.tex
\section{Introduction}
\label{sec:intro}
Question Generation (QG) commonly comprises automatic composition of an appropriate  question given a passage of text and answer located within that text. QG is highly related to the task of machine reading comprehension (MRC), which is a sub-task of question answering (QA)~\cite{du-etal-2017-learning, xie-etal-2020-exploring, pan-etal-2020-semantic, puri-etal-2020-training-synthetic, lyu2021improving, chen2020reinforcement, 10.1007/978-3-030-82136-4_13}. Both QG and MRC receive similar input, a (set of) document(s), while the two tasks diverge on the output they produce, as QG systems generate questions for a predetermined answer within the text while conversely MRC systems aim to answer a prescribed set of questions. 
Recent QG research suggests that the direct employment of MRC datasets for QG tasks is advantageous \citep{qg-model-example1-squad,qg-model-example2-hotpotqa,qg-model-example3-ms-marco}.

In terms of QG evaluation, widely-applied metrics can be categorized into two main classes: word overlap metrics (e.g., BLEU~\cite{papineni-etal-2002-bleu} and Answerability~\cite{metrics-answerability}) and metrics that employ large pre-trained language models (BLEURT~\cite{metric-bleurt} and BERTScore~\cite{metric-bert-score}). 
Evaluation via automatic metrics still face a number of challenges, however.
Firstly, most existing metrics are not specifically designed to evaluate QG systems as they are borrowed from other NLP tasks. 
Since such metrics have been criticized for poor correlation with human assessment in the evaluation of their own NLP tasks such as machine translation (MT) and dialogue systems \cite{review_BLEU, graham2015re, graham2016achieving, ji2022achieving}, thus that raises questions about the validity of results based on such metrics designed for other tasks.
Another challenge lies in the fact that existing automatic evaluation metrics rely on  comparison of a candidate with a ground-truth reference. Such approaches ignore the \textit{one-to-many} nature of QG ignoring the fact that a QG system is capable of generating legitimately plausible questions that will be harshly penalised simply for diverging from ground-truth questions. 
For example, with a passage describing Ireland, the country located in western Europe, two questions $Q1$ and $Q2$, where $Q1$=``\textit{What is the capital of Ireland?}'' and $Q2$=``\textit{Which city in the Leinster province has the largest population?}'', can share the same answer ``\textit{Dublin}''. In other words, it is fairly appropriate for a QG system to generate either $Q1$ or $Q2$ given the same passage and answer, despite few overlap between the meanings of $Q1$ and $Q2$. We deem it the \textit{one-to-many} nature of the QG task, as \textit{one} passage and answer can lead to \textit{many} meaningful questions. A word overlap based metric will however incorrectly assess $Q2$ with a lower score if it takes $Q1$ as the reference because of the lack of word overlap between these two questions. 
A potential solution is to pair each answer with a larger number of hand-crafted reference questions. However, the addition of reliable references requires additional resources, usually incurring a high cost, while attempting to include every possible correct question for a given answer is prohibitively expensive and impractical. Another drawback is that pretrained-model-based metrics require extra resources during the fine-tuning process, resulting in a high cost. Besides the evaluation metrics aforementioned human evaluation is also widely employed in QG tasks. However, the QG community currently lacks a standard human evaluation approach as current QG research employs disparate settings of human evaluation (e.g., expert-based or crowd-sourced, binary or $5$-point rating scales) \cite{xie-etal-2020-exploring, model-hierarchical-seq2seq-and-context-switch}.

To address the existing shortcomings in QG evaluation, we firstly propose a new automatic metric called QAScore. To investigate whether QAScore can outperform existing automatic evaluation metrics, we additionally devise a new human evaluation approach of QG systems and evaluate its reliability in terms of consistent results for QG systems through self-replication experiments. Details of our contributions are listed as follows:

\begin{enumerate}
    \item We propose a pretrained language model based evaluation metric called QAScore, which is unsupervised and reference-free. QAScore utilizes the RoBERTa model \cite{model-roberta}, and evaluates a system-generated question using the cross entropy in terms of the probability that RoBERTa can correctly predict the masked words in the answer to that question.
    
    \item We propose a novel and highly reliable crowd-sourced human evaluation method that can be used as a standard framework for evaluating QG systems. Compared to other human evaluation methods, it is cost-effective and easy to deploy. We further conduct a self-replication experiment showing a correlation of $r=0.955$ in two distinct evaluations of the same set of systems. According to the results of the human evaluation experiment, QAScore can outperform all other metrics without supervision steps or fine-tuning, achieving a strong Pearson correlation with human assessment;
\end{enumerate}

%% file: 2-Related-Work.tex
\section{Background: Question Answering, Question Generation and Evaluation}
\label{sec:background}

\subsection{Question Answering}

Question Answering~(QA) aims to provide answers $a$ to the corresponding questions $q$, Based on the availability of context $c$, QA can be categorized into Open-domain QA~(without context)~\cite{chen2017reading,zhu2021retrieving} and Machine Reading comprehension~(with context)~\cite{rajpurkar-etal-2016-squad1.1, span-extraction-duorc}. Besides, QA can also be categorized into generative QA~\cite{free-answering-narrativeqa,xu-etal-2022-fantastic} and extractive QA~\cite{trischler-etal-2017-newsqa,lyu-etal-2022-extending,lewis2021paq,zhang2020machine}. Generally, the optimization objective of QA models is to maximize the log likelihood of the ground-truth answer $a$ for the given context $c$. Therefore the objective function regarding the parameters $\theta$ of QA models is:

\begin{equation}
    J(\theta) = logP(a|c,q;\theta)
\end{equation}
\subsection{Question Generation}

Question Generation~(QG) is a task where models receive context passages $c$ and answers $a$, then generate the corresponding questions $q$ which are expected to be semantically relevant to the context $c$ and answers $a$~\cite{qg_survey_liangmingpan,lyu2021improving}. Thus QG is a reverse/dual task of QA as QA aims to provide answers $a$ to questions $q$ whereas QG targets at generating questions $q$ for the given answers $a$. Typically, the architecture of QG systems is mainly Seq2Seq model~\cite{NIPS2014_seq2seq} which generates the $q$ word by word in auto-regressive manner. The objective for optimizing the parameters $\theta$ of QG systems is to maximize the likelihood of $P(q|c,a)$:

\begin{equation}
    J(\theta) = logP(q|c,a) = \sum_{i} logP(q_{i}|q_{<i},c,a)
\end{equation}

\subsection{Automatic evaluation metrics}
We introduce two main categories of automatic evaluation metrics applied for QG task: word-overlap-based metrics and pretrained-model-based metrics in the following sections. 

\subsubsection{Word-overlap-based metrics}
Word-overlap-based metrics usually assess the quality of a QG system according to the overlap rate between the words of a system-generated candidate and a reference. Most of such metrics, including BLEU, GLEU, ROUGE and METEOR are initially proposed for other NLP tasks (e.g., BLEU is for MT and ROUGE is for text summarization), while Answerability is a QG-exclusive evaluation metric. 
\begin{flushleft}
\textbf{BLEU} Bilingual Evaluation Understudy (BLEU) is a method that is originally proposed for evaluating the quality of MT systems \citep{metrics-bleu}. For QG evaluation, BLEU computes the level of correspondence between a system-generated question and the reference question by calculating the precision according to the number of $n$-gram matching segments. These matching segments are thought to be unrelated to their positions in the entire context. The more matching segments there are, the better the quality of the candidate is. 
\end{flushleft}




\begin{flushleft}
\textbf{GLEU} GLEU (Google-BLEU) is proposed to overcome the drawbacks of evaluating a single sentence \citep{metrics-gleu}. As a variation of BLEU, the GLEU score is reported to be highly correlated with the BLEU score on a corpus level. GLEU uses the scores of precision and recall instead of the modified precision in BLEU. 
\end{flushleft}


\begin{flushleft}
\textbf{ROUGE} Recall-Oriented Understudy for Gisting Evaluation (ROUGE) is an evaluation metric developed for the assessment of the text summarization task, but originally adapted as a recall-adaptation of BLEU \citep{metrics-rouge}. ROUGE-L is the most popular variant of ROUGE, where L denotes the longest common subsequence (LCS). The definition of LCS is a sequence of words that appear in the same order in both sentences. In contrast with sub-strings (e.g., $n$-gram), the positions of words in a sub-sequence are not required to be consecutive in the original sentence. ROUGE-L is then computed by the F-$\beta$ score according to the number of words in the LCS between a question and a reference.
\end{flushleft}




\begin{flushleft}
\textbf{METEOR} Metric for Evaluation of Translation with Explicit ORdering (METEOR) was firstly proposed to make up for the disadvantages of BLEU, such as lack of recall and the inaccuracy of assessing a single sentence \citep{metrics-meteor-1.0}. METEOR first generates a set of mappings between the question $q$ and the reference $r$ according to a set of stages, including: exact token matching (i.e., two tokens are the same), WordNet synonyms (e.g., \textit{well} and \textit{good}), and Porter stemmer (e.g., \textit{friend} and \textit{friends}). METOER score is then computed by the weighted harmonic mean of precision and recall in terms of the number of unigrams in mappings between a question and a reference.
\end{flushleft}



\begin{flushleft}
\textbf{Answerability} Aside from the aforementioned evaluation methods - which are borrowed from other NLP tasks, an automatic metric called Answerability is specifically proposed for the QG task \citep{metrics-answerability}. \citet{metrics-answerability} suggest combining it with other existing metrics since its aim is to measure how answerable a question is, something not usually targeted by other automatic metrics. For example, given a reference question $r$: ``\textit{What is the address of DCU?}'' and two generated questions $q_1$:  ``\textit{address of DCU}'' and $q_2$: ``\textit{What is the address of}'', it is obvious that $q_1$ is rather answerable since it contains enough information while $q_2$ is very confusing. However, any similarity-based metric is certainly prone to think that $q_2$ (ROUGE-L: $90.9$; METEOR: $41.4$; BLEU-1: $81.9$) is closer to $r$ than $q_1$ (ROUGE-L: $66.7$; METEOR: $38.0$; BLEU-1: $36.8$). Thus, Answerability is proposed to solve such an issue. In detail, for a system-generated question $q$ and a reference question $r$, the Answerability score can be computed as shown in Equation \ref{eq:answerability}:
\end{flushleft}

\begin{equation}
\label{eq:answerability}
\begin{aligned}
    P &= \sum_{i\in E}w_i\dfrac{h_i(q,r)}{k_i(q)} \\
    R &= \sum_{i\in E}w_i\dfrac{h_i(q,r)}{k_i(r)} \\
    Answerability &= \dfrac{2\times P\times R}{P+R}
\end{aligned}
\end{equation}
where $i$ ($i \in E$) represents certain types of elements in $E=\{R,N,Q,F\}$ ($R=$ Relevant Content Word, $N=$ Named Entity, $Q=$ Question Type, and $F=$ Function Word).
$w_i$ is the weight for type $i$ that ${\displaystyle \sum\nolimits_{i\in E}w_i = 1 }$. Function $h_i(x,y)$ returns the number of $i$-type words in question $x$ that have matching $i$-type words in question $y$, and $k_i(x)$ returns the number of $i$-type words occuring in question $x$. The final Answerability score is the F1 score of Precision $P$ and Recall $R$. 

Along with using Answerability individually, a common practice is to combine it with other metrics as suggested when evaluating QG systems \cite{qblue-use1,qblue-use2}:
\begin{multline}
    \label{eq:answerability-with-metric}
    Metric_{mod} = \beta\cdot Answerability + \\(1-\beta)\cdot Metric_{ori}
\end{multline}
where $Metric_{mod}$ is a modified version of an original evaluation metric $ Metric_{ori}$ using Answerability, and $\beta$ is a hyper-parameter. In this experiment, we combine it with BLEU to generate the $Q$-BLEU score using the default value of $\beta$.

\subsubsection{Pretrained-model-based metrics}

\begin{flushleft}
\textbf{BERTScore} \citet{metric-bert-score} proposed an automatic metric called BERTScore for evaluating text generation task because word-overlap-based metrics like BLEU fail to account for compositional diversity.
Instead, BERTScore computes a similarity score between tokens in a candidate sentence and its reference based on their contextualized representations produced by BERT \citep{model-bert}. Given a question that has $m$ tokens and a question that has $n$ tokens, the BERT model can first generate the representations of $q$ and $r$ as $q = \langle q_1,q_2,\dotsc, q_m \rangle$ and $r = \langle r_1,r_2,\dotsc, r_n \rangle$, where $q_i$ and $r_i$ respectively mean the contextual embeddings of the $i$-th token in $q$ and $r$. Then, the BERT score between the question and the reference can be computed by Equation \ref{eq:qg-bertscore}:
\end{flushleft}
\begin{equation}
\label{eq:qg-bertscore}
\begin{split}
P_{\text{BERT}} &= \frac{1}{m}\sum_{p_i\in p}\operatorname*{max}_{r_j \in r} p_i^{\top} r_j\\
R_{\text{BERT}} &= \frac{1}{n}\sum_{r_i\in r}\operatorname*{max}_{p_j \in p} p_j^{\top} r_i\\
\text{BERTScore} &= \frac{2\cdot P_{\text{BERT}} \cdot R_{\text{BERT}} }{ P_{\text{BERT}} + R_{\text{BERT}} }
\end{split}
\end{equation}
where the final BERTScore is the F1 measure computed by precision $P_{\text{BERT}}$ and recall $R_{\text{BERT}}$.

\begin{flushleft}
\textbf{BLEURT} BLEURT is proposed to solve the issue that metrics like BLEU may correlate poorly with human judgments \cite{metric-bleurt}. It is a trained evaluation metric which takes a candidate and its reference as input and gives a score to indicate how the candidate can cover the meaning of the reference.
BLEURT uses a BERT-based regression model trained on the human rating data from the WMT Metrics Shared Task from 2017 to 2019. Since BLEURT was proposed for evaluating models on the sentence level, meanwhile no formal experiments are available for corpus-level evaluation, we directly compute the final BLEURT score of a QG system as the arithmetic mean of all sentence-level BLEURT scores in our QG evaluation experiment as suggested (see the discussion on \url{https://github.com/google-research/bleurt/issues/10}).
\end{flushleft}
\subsection{Human evaluation}
\label{sec:background-sub-human-eval}
Although the aforementioned prevailing automatic metrics mentioned above are widely employed for QG evaluation, criticism of $n$-gram overlap-based metrics' ability to accurately and comprehensively evaluate the quality has also been highlighted \citep{qg-bleu-doubt}. As a single answer can potentially have a large number of corresponding plausible questions, simply computing the overlap rate between an output and a reference to reflect the real quality of a QG system does not seem convincing. A possible solution is to obtain more correct questions per answer, as $n$-gram overlap-based metrics would usually benefit from multiple ground-truth references. However, this may elicit new issues: 1) adding additional references over the entire corpora requires similar effort to creating a new data set incurring expensive and time resource costs; 2) it is not straightforward to formulate how word overlap should contribute to the final score for systems. 

Hence, human evaluation is also involved when evaluating newly proposed QG systems. A common approach is to evaluate a set of system-generated questions and ask human raters to score these questions on an $n$-point Likert scale. Below we introduce and describe recent human evaluations are applied to evaluate QG systems.

\citet{qg-model-n-point-example1} proposed EQG-RACE to generate examination-type questions for educational purposes. 100 outputs are sampled and three expert raters are required to score these outputs in three dimensions: fluency - \textit{whether a question is grammatical and fluent}; relevancy - \textit{whether the question is semantically relevant to the passage}; and answerability - \textit{whether the question can be answered by the right answer}. A 3-point scale is used for each aspect, and aspects are reported separately without overall performance.

KD-QG is a framework with a knowledge base for generating various questions as a means of data augmentation \citep{qg-model-n-point-example2}. For its human evaluation, three proficient experts are individually assigned to 50 randomly-sampled items and judge whether an assigned item is \textit{reliable} on a binary scale ($0$-$1$). Any item with a positive reliability will be further assessed for its level of \textit{plausibility} on a $3$-point scale ($0$-$2$) that is construed as: $0$ - \textit{obviously wrong}, 1 - \textit{somewhat plausible} and $2$ - \textit{plausible}. These two aspects are treated separately without reporting any overall ranking.

Answer-Clue-Style-aware Question Generation (ACS-QG) aims to generate questions together with the answers from unlabeled textual content \citep{qg-model-n-point-example3}. Instead of evaluating the questions alone, a sample is a tuple of $(p,q,a)$ where $p=$ passage, $q=$ question and $a=$ answer. A total of 500 shuffled samples are assigned to ten volunteers, where each volunteer receives 150 samples to ensure an individual sample is evaluated by three different volunteers. 
Three facets of a sample are evaluated: \textit{well-formedness} (yes/understandable/no) - \textit{if the question is well-formed}; \textit{relevancy} (yes/no) - \textit{if the question is relevant to the passage}; and \textit{correctness} (yes/partially/no) - \textit{if the answer the question is correct}. The results for each facet are reported as percentages rather than as a summarized score. 

\citet{qg-model-n-point-example4} proposed a neural QG model consisting of two mechanisms: semantic matching and position inferring. The model is evaluated by human raters for three aspects: \textit{semantic-matching}, \textit{fluency}, and \textit{syntactic-correctness} on a $5$-point scale. However, the details about: 1) the number of evaluated samples; 2) the number of involved raters; 3) the type of human raters (crowd-sourced or experts) are unfortunately not provided. 

QURIOUS is a pretraining method for QG, and QURIOUS-based models are expected to outperform other non-QURIOUS models \citep{qg-model-n-point-example5}. To verify this, a crowd-sourced human evaluation experiment is then conducted. Thirty passages with answers are randomly selected, and human raters  compare questions from two distinct models. For each single comparison, 3 individuals are involved. Specifically, a human rater is presented with a passage, an answer, and questions $A$ and $B$ from two models, and is asked to rate which question is better than the other according to two aspects: \textit{naturalness} - the question is fluent and written in well-formed English, and correctness - \textit{the question is correct given the passage and the answer}. Each comparison has one of the three distinct choices: $(A=best, B=worst)$, $(A=equal, B=equal)$ and $(A=worst, B=best)$, and the final human score of a system for each aspect is computed as the number of times it is rated as $best$ subtracting the number of times it is rated as $worst$, followed by dividing by the number of times it is evaluated in total.

Although the fact that human evaluation is somewhat prevalent in the QG evaluation much more than many other NLP areas, there still remains three major issues:
\begin{enumerate}

    \item There still lacks a standard human evaluation for QG since the aforementioned examples individually use disparate rating options and settings with only a few overlaps. These existing methods for the QG task can generally change from one set of experiments to the next, highlighting the lack of a standard approach, making comparisons challenging;
    \item The vast majority of QG human evaluation methods are either expert-based or volunteer-based, with the former are normally expensive and latter likely incurring issues such as shortages of rater availability. Furthermore, the inconvenience of deploying human evaluation at scale can lead to a small sample size, which could possibly hinder the reliability of evaluation results;
    \item Much of the time, details of human evaluation experiments are vague with on occasion sample sizes and number of raters omitted from publications. Although expert-based human evaluation can be deemed to have a high level of rater's agreement, such information is seldom reported, resulting in difficulties interpreting the reliability and validity of experiments, in particular when crowd-sourced human evaluation is employed.
\end{enumerate}

%% file: 3-Proposed-Metric.tex
\section{QAScore - An Automatic Metric for Evaluating QG systems using Cross-Entropy}




Since QG systems are required to generate a question according to a passage and answer, we think that the evaluation of the question should take into account the passage and answer as well, which current metrics fail to achieve.
In addition, the QG evaluation should consider the \textit{one-to-many} nature (see Section \ref{sec:intro}) that there may be many appropriate questions based on one passage and answer, while current metrics usually have only one or a few references to refer to. Furthermore, metrics, such as BERTScore and BLEURT, require extra resources for fine-tuning, which is expensive and inconvenient for utilization.
Hence, we propose a new automatic metric, which has three main advantages compared with exsiting automatic QG evaluation metrics.: 1) it can directly evaluate a candidate question with no need to compute the similarity with any human-generated reference; 2) it is easy to deploy as it takes a pretrained language model as the scorer and requires no extra data for further fine-tuning; 3) it can correlate better with humans according to human judgements than other existing metrics. 
\input{subsections/05-lm-qg-score}

%% file: subsections/05-lm-qg-score.tex
\subsection{Proposed metric}
In this section, we describe our proposed pretrained-model-based unsupervised QG metric. The fact that there are many possible correct questions for the same answer and passage means that multiple distinct questions can legitimately share the same answer, due to the one-to-many nature of QG task as described in Section \ref{sec:qg-experiment-design}. Hence, we think a reference-free metric is more appropriate since there can be several correct questions for a given pair of an answer and a passage. In addition, the QG and QA tasks are complementary where the common practice of QG is to generate more data to augment a QA dataset and QA systems can benefit from the augmented QA dataset \cite{qg-data-aug-for-qa1}. Therefore, we believe a QA system should be capable of judging the quality of the output of a QG system, and the proposed metric is designed to score a QG output in a QA manner whose detail will be introduced in Section \ref{qascore-subsec-methodology}.
Furthermore, our proposed metric has the advantage of being unsupervised. Pretrained language models are demonstrated to contain plenty of useful knowledge since they are trained on large scale corpus \cite{roberta-ability}. Therefore, we plan to directly employ a pretrained language model to act as a evaluation metric without using other training data or supervision, as introduced in Section \ref{qascore-subsec-roberta}.

%

\subsection{Methodology}
\label{qascore-subsec-methodology}

Since QG and QA are two complementary tasks, we can naturally conjecture that a QG-system-generated question can be evaluated according to the quality of the answer generated by a QA system. Therefore, we think the likelihood of the generated question $q$ to the given answer $a$ and passage $p$ should be proportional to the likelihood of the corresponding answer $a$ to the generated question $q$ and passage $p$:
\begin{equation}
    P(q|p,a) \propto P(a|p,q)
\end{equation}

We take the passage and the answer $a$, ``commander of the American Expeditionary Force (AEF) on the Western Front'', in Figure  \ref{tbl:qg-instruction-para-and-question} as an example. We show two distinct question $q_1$ and $q_2$, where $q_1$ is ``\textit{What was the most famous post of the man who commanded American and French troops against German positions during the Battle of Saint-Mihiel?}'' and $q_2$ is ``\textit{What was the Battle of Saint-Mihiel?}''. It can be found that, $a$ is the correct answer to $q_1$ rather than $q_2$. Therefore, in this case a QA model is \textit{more} likely to generate $a$ when given $q_1$, and it is expected \textit{not} to generate $a$ when given $q_2$. In another words, the likelihood that a QA model can produce $a$ given $q_1$ is more than that given $q_2$:

\begin{equation}
    P(q_1|p, a) > P(q_2|p,a)
\end{equation}

The detailed scoring mechanism will be introduced in Section \ref{qascore-subsec-process-of-scoring}.



\subsubsection{Pre-trained Language Model - RoBERTa}
\label{qascore-subsec-roberta}
We chose to employ the masked language model RoBERTa \cite{model-roberta} in a MRC manner to examine the likelihood of an answer, and its value can act as the quality of the target question to be evaluated. RoBERTa (\textbf{R}obustly \textbf{o}ptimized \textbf{BERT} \textbf{a}pproach) is a BERT-based approach for pretraining a masked language model. Compared with the original BERT, RoBERTa is trained on a larger dataset with a larger batch size and longer elapsed time. It also removes the next sentence prediction (NSP) step and leverages full-sentences (sentences that reach the maximal length). For text encoding, RoBERTa employs a smaller BPE (Byte-Pair Encoding) vocabulary from GPT2 instead of the character-level BPE vocabulary used in the original BERT. 


In general, the pre-training objective of RoBERTa aims to predict the original tokens which are replaced by a special token \cite{model-roberta}. Given a sequence $(w_1, w_2, \dots ,w_n)$, a token $w$ in the original sentence is randomly replaced by a special token \textit{[MASK]}. And the pre-training objective of RoBERTa can be formulated as Equation \ref{eq:roberta-pretrain}:

\begin{multline}
    \label{eq:roberta-pretrain}
        J(\theta) = log P(\hat W|\tilde{W}) = \sum_{i\in \hat I} logP(w_{i}|w_{j_1}, \\ w_{j_2},......, w_{j_n};j_{k} \in \{I-\hat I\})
\end{multline}

where $\hat W$ and $\tilde{W}$ represent masked words and unmasked words respectively, $I$ denotes the original indices of all tokens including masked and unmasked tokens, $\hat I$ represents the indices of masked tokens, and the indices of unmasked tokens can be  denoted as $I-\hat I$. 

\subsubsection{Process of scoring}
\label{qascore-subsec-process-of-scoring}
\begin{figure*}[t]
    \centering
    \includegraphics[width=\textwidth]{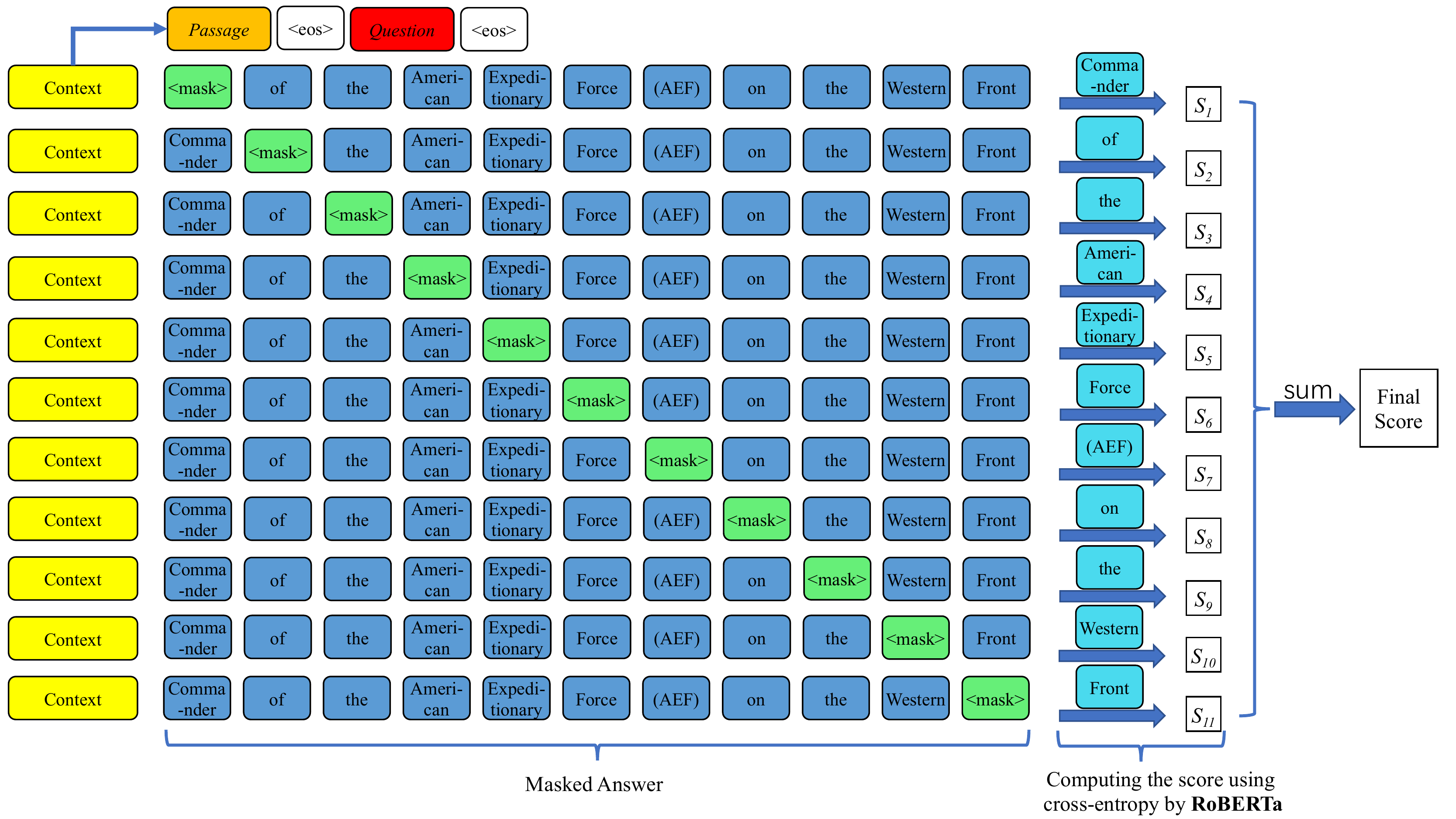}
    \caption{The process of scoring a question by RoBERTa, where the context (yellow) contains the passage and the question (to be evaluated), $\langle$eos$\rangle$ is the separator token, the score of a single word is the likelihood that RoBERTa can predict the real word (cyan) which is replaced by the mask token $\langle$mask$\rangle$ (green) in the original answer, and the final metric score is the sum of scores of all words in the answer.}
    \label{fig:qg-lm-scoring-process}
\end{figure*}

Since this proposed metric leverages a means of question answering to assess QG-system-generated questions, we call it QAScore. Given the passage, the correct answer, and the QG-system-generated question, we first encode and concatenate the passage and the answer. Figure \ref{fig:qg-lm-scoring-process} provides a visualization of the process of scoring a generated question using its passage and answer using the masked language model RoBERTa. First, the passage and the question are concatenated by the end-of-sequence token $\langle$eos$\rangle$, which represents the context for the masked language model. Next, the masked answer containing one masked word is concatenated by the context together with the $\langle$eos$\rangle$ token as the input for the model. The model is then asked to predict the real value of the masked word using the context and the masked answer. The likelihood that RoBERTa can generate the true word can act as the score for that masked word. 
For the evaluation of a single question, all words in the given answer will be masked in a one-at-a-time manner.
The final metric score of the question $Q$ can be computed by the following Equation:
\begin{align}
    o_{w} &= M(P,Q,A_{\widetilde{w}}) \label{eq:01} \\
    p_{w} &= log \frac{e^{o_{w}}}{\sum_{w'}{e^{o_{w'}}}} \label{eq:02}\\
    l_{w} &= \sum_{c=1}^{C}{y_{w,c}\cdot p_{w,c}} \label{eq:03} \\
    QAScore(Q) &= \sum_{w \in {A}}{l_{w}} \label{eq:04}
\end{align}
where Equation \ref{eq:01} computes the output of the model RoBERTa ($M$) when receiving the passage $P$, the question $Q$, and $A_{\widetilde{w}}$ which is the Answer $A$ with a word $w$ in it masked, as the input. Then, Equation \ref{eq:02} computes the probabilities of $o_{w}$ using the log-softmax function. Equation \ref{eq:03} is the log likelihood of $p_{w}$ where $C$ is the number of vocabulary size of RoBERTa. Finally, the QAScore of the question $Q$ is the sum of $l_{w}$ among all words in its relevant Answer $A$.

\input{subsections/03-dataset-and-system}

\subsection{Results}

\begin{table*}[ht]
    \centering
    \caption{The scores of all QG systems, including the overall human scores ($z$), this proposed evaluation metric QAScore, and other existing metrics, where QG systems are sorted according to human scores.}
    \label{tbl:qg-lm-score}

\input{tables/tbl-qg-autometric}

\end{table*}

Table \ref{tbl:qg-lm-score} shows the human scores ($z$) and the metric scores of QG systems evaluated using QAScore and current prevailing QG evaluation metrics, where the human score is computed according to our newly proposed human evaluation method, which will be introduced will be introduced in Section \ref{sec:human-eval}. Table \ref{tbl:qg-automatic-corr} describes how these metrics correlate with human judgements according to the results of the human evaluation experiment. Since our metric does not rely on a ground-truth reference, we can additionally include the result of the Human system unlike other automatic metrics. It can be seen that our metric correlates with human judgements at $0.864$ according to the Pearson correlation coefficient, where even the best automatic metric METEOR can only reach $0.801$ (see Table \ref{tbl:qg-automatic-corr}). Also, compared with the other two pretrained-model-based metrics BERTScore and BLEURT, our metric can outperform them at $>0.1$. In terms of Spearman, our metric achieves $\rho \approx 0.8$ where other metrics can only reach at most $\rho \approx 0.6$. In addition, our metric also outperforms other metrics according to Kendall’s tau
since it reach at $\tau \approx 0.7$ and other metrics merely achieve at most $\tau \approx 0.5$. We can conclude that our metric correlates better with human judgements with respect to all three categories of correlation coefficients.

\begin{table*}[t]
    \centering
    \caption{The Pearson ($r$), Spearman ($\rho$) and Kendall's tau ($\tau$) correlation between automatic metric scores and human judgements according to the overall scores in the first run, where the metrics are sorted by $r$.}
    \input{tables/tbl-qg-autometric-corr}
    \label{tbl:qg-automatic-corr}
\end{table*}

%% file: subsections/03-dataset-and-system.tex
\subsection{Dataset and QG systems}
\label{sec:qg-dataset-and-systems}

\subsubsection{HotpotQA dataset}
We conduct the experiment on the HotpotQA dataset \citep{qg-hotpotqa}, initially proposed for the multi-hop question answering task (see \url{https://hotpotqa.github.io/}).
The term, multi-hop, means that a machine should have the ability to answer given questions by extracting useful information from several related passages. 
The documents in the dataset are extracted from Wikipedia articles, and the questions and answers are created by crowd workers. A worker is asked to provide the questions whose answers requires reasoning over all given documents.
Each question in the dataset is associated with one correct answer and multiple passages, where the answer is either a sub-sequence from the passage or simply yes-or-no.
These multiple passages are treated as a simple passage to show to human raters during the experiment. Note that the original HotpotQA test set provides no answer for each question, and such a set is inappropriate for the QG task as an answer is necessary for a QG system to generate a question. Instead, a common practice is to randomly sample a fraction from the training set as the validation set, and the original validation set can act as the test set when training or evaluating a QG system based on a QA dataset. The test set we used to grab system-generated outputs for the QG evaluation is in fact the validation set. 

Besides, HotpotQA dataset provides two forms of passages: full passages and supporting facts. For each question, its full passages, on the average, consist of 41 sentences while the average number of sentences in its supporting facts is 8. 
Since the reading quantity is one of our concerns, we use the sentences from supporting facts to constitute the passage to prevent workers from reading too many sentences per assignment.

\subsection{QG systems for evaluation}
\label{subsec:qg-systems}
To analyze the performance of our proposed evaluation method, 11 systems will be evaluated, including 10 systems that are trained on the HotpotQA dataset and the Human system that can represent the performance of humans on generating questions. The Human system is directly made up of the questions extracted from the HotpotQA testset. The 10 trained systems are from the following neural network models:
\begin{itemize}
    \item \textbf{T5 (small \& base)}: a model using a text-to-text transfer transformer that is pretrained on a large text corpus \cite{model-t5};
    \item \textbf{BART (base \& large)}: a denoising auto-encoder using the standard sequence-to-sequence transformer architecture \cite{model-bart};
    
    \item \textbf{Att-GGNN}: an attention-based gated graph neural network model \citep{model-Att-GGNN};
    
    \item \textbf{Att-GGNN (plus)}: a variant of Att-GGNN model which is combined with the context switch mechanism \cite{model-hierarchical-seq2seq-and-context-switch};
    \item \textbf{H-Seq2seq}: a hierarchical encoding-decoding model proposed for the QG task \cite{model-hierarchical-seq2seq-and-context-switch};
    
    \item \textbf{H-Seq2seq$^*$}: a variant of H-Seq2seq which utilizes a larger dictionary for the avoidance of generating the unknown token $\langle \text{UNK} \rangle$;
    
    \item \textbf{GPT-2}: a large transformer-based language model with parameters reaching the size of 1.5B \cite{model-gpt-2}.
    
    \item \textbf{RNN}: a sequence-to-sequence model using the vanilla current neural network (RNN) structure \cite{model-rnn}.
\end{itemize}

These systems then generate questions on the HotpotQA testset.

%% file: tables/tbl-qg-autometric.tex
\begin{tabularx}{\linewidth}{lYYYYYYYY}
\hline
\multicolumn{1}{c}{System} & $z$ &  \rotatebox{45}{QAScore} & \rotatebox{45}{METEOR} & \rotatebox{45}{ROUGE-L} & \rotatebox{45}{BERTScore}  & \rotatebox{45}{BLEURT} & \rotatebox{45}{$Q$-BLEU4} & \rotatebox{45}{$Q$-BLEU1} \bigstrut\\
\hline
Human & \phantom{$-$}0.322 & $-$0.985 &  --  & --  & --  & --  & --  & -- \bigstrut[t]\\
BART$_{large}$ & \phantom{$-$}0.308 & $-$1.020 &  30.18  & 47.58  & 90.85  & $-$0.363  & 43.77  & 51.47 \\
BART$_{base}$ & \phantom{$-$}0.290 & $-$1.030 &  29.66  & 47.13  & 90.74  & $-$0.381  & 44.14  & 51.65  \\
T5$_{base}$ & \phantom{$-$}0.226 & $-$1.037 &  27.99  & 41.60  & 88.44  & $-$0.682  & 37.78  & 44.84  \\
RNN   & \phantom{$-$}0.147 & $-$1.064 &  15.46  & 26.77  & 84.59  & $-$1.019  & \phantom{0}9.68  & 15.92  \\
H-Seq2seq & \phantom{$-$}0.120 & $-$1.076 &  17.50  & 29.86  & 85.49  & $-$0.953  & 10.51  & 17.74  \\
T5$_{small}$ & \phantom{$-$}0.117 & $-$1.049 &  23.62  & 32.37  & 86.34  & $-$0.860  & 26.73  & 32.92  \\
Att-GGNN$_{plus}$ & \phantom{$-$}0.076 & $-$1.065  &  21.77  & 36.31  & 86.27  & $-$0.784  & 12.63  & 19.86  \\
H-Seq2seq$^*$ & \phantom{$-$}0.053  & $-$1.045 &  18.23  & 31.69  & 85.83  & $-$0.866  & 11.12  & 18.36  \\
Att-GGNN & $-$0.008 & $-$1.068  &  20.02  & 33.60  & 86.00  & $-$0.802  & 11.13  & 18.67  \\
GPT-2 & $-$0.052  & $-$1.108  &   16.40  & 29.98  & 86.44  & $-$0.899  & 24.83  & 31.85  \bigstrut[b]\\
\hline
\end{tabularx}

%% file: tables/tbl-qg-autometric-corr.tex


\begin{tabularx}{0.9\linewidth}{cYYYYYYY}
\hline
& \rotatebox{45}{QAScore} & \rotatebox{45}{METEOR} & \rotatebox{45}{ROUGE-L} & \rotatebox{45}{BERTScore}  & \rotatebox{45}{BLEURT} & \rotatebox{45}{$Q$-BLEU4} & \rotatebox{45}{$Q$-BLEU1} \bigstrut\\
\hline
$r$ & 0.864 & 0.801  & 0.770  & 0.761  & 0.739  & 0.725  & 0.724  \bigstrut[t]\\
$\rho$ & 0.827 & 0.612  & 0.503  & 0.430  & 0.503  & 0.467  & 0.467  \\
$\tau$ & 0.709 & 0.511  & 0.378  & 0.289  & 0.378  & 0.289  & 0.289  \bigstrut[b]\\
\hline
\end{tabularx}

%% file: 4-Proposed-Human-Evaluation.tex
\section{New Human Evaluation Methodology}
\label{sec:human-eval}


To investigate the performances of QAScore and current QG evaluation metrics, and to overcome the issues described in Section \ref{sec:background}, we propose a new human evaluation method for assessing QG systems in this section. First, this can be used as a standard framework for evaluating QG systems because of its flexible evaluation criteria unlike other model-specific evaluation methods. Second, it is a crowd-sourcing human evaluation rather than expert-based, thus it can be deployed on a large scale within an affordable budget. Furthermore, the self-replication experiment proves the robustness of our method, and we provide the specified details of our method and corresponding experiment for reproduction and future studies.

\input{subsections/01-exp-design}

\input{subsections/02-construction}

%% file: subsections/01-exp-design.tex
\subsection{Experiment design}
\label{sec:qg-experiment-design}
In this section, the methodology of our proposed crowd-sourcing human evaluation for QG is introduced. An experiment that investigates the reliability of results for the new method is also provided, as well as details such as the design of interface shown to human raters, mechanisms for  quality checking the evaluation, and the evaluation criteria employed.

\subsubsection{Methodology}

QG receives a context with a sentence as the input and generates a textual sequence as the output, with automatic metrics reporting the computation of word/$n$-gram overlap between the generated sequence and the reference question. However, human evaluation can vary. When evaluating MRC systems via crowd-sourced human evaluation, raters are asked to judge system-generated answers with reference to gold standard answers because a correct answer to the given question should be, to some degree, similar to the reference answer \cite{mrc-human-eval}. 

Whereas, simply applying the same evaluation is not ideal since evaluating a QG system is more challenging due to its \textit{one-to-many} nature (see Section \ref{sec:intro}), namely a QG system can produce a question that is appropriate but distinct from the reference. Such evaluation may unfairly underrate the generated question because of its inconformity with the reference. To avoid this situation in our experiment, we ask a human rater to directly judge the quality of a system-generated question with its passage and answer present, instead of a reference question.



\subsubsection{Experiment user interface}

Since our crowd-sourced evaluation method can involve workers who have no specific knowledge of the related field, a minimal level of guidance is necessary to concisely introduce the evaluation task. Prior to each HIT, a list of instructions followed by button labelled  \textit{I understand} is provided, with the human rater beginning a HIT by clicking the button. The full list of instructions is described in Figure \ref{tbl:qg-instruction}. In regard to the fourth instruction, Chrome browser is recommended to ensure the stability because we present the HTML element ``range control'' embedded with hashtags while not all browsers can fully support this feature (e.g., Firefox does not support this feature at all).
\begin{figure}[ht]
 \centering
 \input{tables/tbl-qg-instruction}
 \caption{Full instructions shown to a crowd-sourced human assessor read prior to starting HITs}
  \label{tbl:qg-instruction}
\end{figure}

Within each HIT, a human assessor is required to read a passage and a system-generated question with the input (correct) answer, then rate the quality of the question according to the given passage and the answer. Since the answer is a sub-sequence of the passage, we directly emphasize the answer within the passage.
Figure \ref{tbl:qg-instruction-para-and-question} provides an example of the interface employed in  experiments, where a worker is shown a passage whose highlighted contents are expected to be the answer to the generated question. Meanwhile, workers may see a passage without any highlighted content since a fraction of the answers are simply ``yes-or-no''.
\begin{figure*}[ht]
 \centering
 \input{tables/tbl-qg-interface-para-and-question}
 \caption{The interface shown to human workers, including a passage with highlighted contents and a system-generated question. The worker is then asked to rate the question.}
  \label{tbl:qg-instruction-para-and-question}
\end{figure*}

\subsubsection{Evaluation criteria}
\begin{figure*}[ht]
    \centering
    \includegraphics[width=\linewidth]{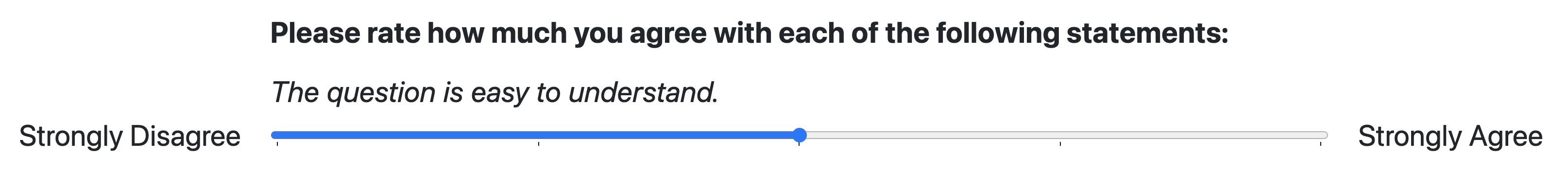}
    \caption{The example of a Likert statement of an evaluation criterion shown to a human worker.}
    \label{fig:qg-interface-likert-statement}
\end{figure*}

Human raters assess the system output question in regards to a range of different aspects (as opposed to providing a single overall score). Figure \ref{fig:qg-interface-likert-statement} provides an example rating criterion, where a human rater is shown a Likert statement and asked to indicate the level of agreement with it through a range slider from \textit{strongly disagree} (left) to \textit{strongly agree} (right). 

The full list of evaluation criteria we employed in this experiment is available in Table \ref{tbl:qg-rate-labels}, where the labels are in reality not shown to the workers during the evaluation. As an empirical evaluation method, these criteria are those most commonly employed in current research but can be substituted for distinct criteria if needed (see Section \ref{sec:background-sub-human-eval}).
Since our contribution focuses on proposing a human evaluation approach that can act as a standard for judging QG systems, rather than proposing a fixed combination of evaluation criteria, the criteria we employed are neither immutable nor hard-coded. And we encourage adjusting, extending and pruning them if necessary. Additionally, the rating criterion ``answerability'' in Table \ref{tbl:qg-rate-labels} should not be confused with the automatic metric Answerability.
\begin{table*}[ht]
    \centering
    \caption{The rating criteria of assessing the quality of a system-generated question. Note that only the Likert statements are available for human workers and the labels are not shown in the experiment.}
    \input{tables/tbl-qg-rate-labels}
    \label{tbl:qg-rate-labels}
\end{table*}

%% file: tables/tbl-qg-instruction.tex
\begin{tabularx}{1.0\linewidth}{rX}
\hline
1. & Each time you will see a question together with a passage, and your task is to rate the questions after reading the given passage.  \bigstrut[t] \\
2. & Each HIT contains one certain passage with 20 various questions to rate. \\
3. & The highlighted content in the passage is expected to be the answer to the presented question. \\
  & A passage with no highlighted content means the question should be a "yes-or-no" question.  \\
4. & Chrome is preferred, other browsers may cause some errors.  \\
5. & There is a feedback box at the end of the HIT. If you encounter any problems, please enter them in this box or email our MTurk account.  \bigstrut[b] \\

\hline
\end{tabularx}

%% file: tables/tbl-qg-interface-para-and-question.tex
\begin{tabularx}{1.0\linewidth}{rX}


\textbf{Passage:} & The Battle of Saint-Mihiel was a major World War I battle fought from 12–15 September 1918 , involving the American Expeditionary Force (AEF) and 110,000 French troops under the command of General John J. Pershing of the United States against German positions . The U.S. Army Air Service ( which later became the U.S. Air Force ) played a significant role in this action . General of the Armies John Joseph ``Black Jack'' Pershing ( September 13 , 1860 – July 15 , 1948 ) was a senior United States Army officer . His most famous post was when he served as the \underline{\textbf{commander of the American Expeditionary Force (AEF) on the Western Front}} in World War I , 1917–18. \\
 & \\
\textbf{Question:} & What was the most famous post of the man who commanded American and French troops against German positions during the Battle of Saint-Mihiel ? \\

\end{tabularx}

%% file: tables/tbl-qg-rate-labels.tex
\begin{tabularx}{1.0\linewidth}{lX}
\hline
\multicolumn{1}{c}{Label} & \multicolumn{1}{c}{Likert statement} \bigstrut\\
\hline
\emph{\textbf{Understandability}}   & \emph{The question is easy to understand.}    \bigstrut[t]\\
\emph{\textbf{Relevancy}} & \emph{The question is highly relevant to the content of the passage.}  \\
\emph{\textbf{Answerability}}      & \emph{The question can be fully answered by the passage}   \\
\emph{\textbf{Appropriateness}} & \emph{The question word (where, when, how, etc.) is fully appropriate.}   \bigstrut[b]\\

\hline
\end{tabularx}

%% file: subsections/02-construction.tex
\subsection{Quality control}
\label{exp-design-subsec-qc}
Similar to human evaluation experiments in other tasks (e.g., MT and MRC) \cite{da}, quality-controlling the crowd-sourced workers is likewise necessary for the QG evaluation. Since no ground-truth reference will be provided for the comparison with system-generated questions, the quality control methods involve no ``reference question''. 
Two methods - \textit{bad reference} and \textit{repeat} - are employed the means of quality-controlling the crowd to filter out incompetent results.

\textbf{Bad reference}: A set of system-generated questions are randomly selected, and their degraded versions are automatically generated to make a set of bad references. To create a \textit{bad reference} question, we took the original system-generated question and degraded its quality by replacing a random short sub-string from it with another string. The replacement samples are extracted from the entire set passages and should have the same length with the replaced string. Given the original question that consists of $n$ words, the number of words that the replacement should have is subsequently decided on the following rules:
\begin{itemize}
    \item for $1\leq n\leq 3$, it comprises $1$ word.
    \item for $4\leq n\leq 4$, it comprises $2$ words.
    \item for $6\leq n\leq 8$, it comprises $3$ words.
    \item for $9\leq n\leq 15$, it comprises $4$ words.
    \item for $16\leq n\leq 20$, it comprises $5$ words.
    \item for $n\geq 21$,  it comprises $\lfloor n/5 \rfloor$ words.
\end{itemize}
Initial and final words are not included for questions with more than two words, and the passage regarding the current question is also excluded.

\textbf{Repeat}: a set of system-generated questions are randomly selected, and they are copied to make a set of repeats. 

In order to implement quality control, we will apply a significance test between the paired bad references and their associate ordinary questions on all rating types. In this case, a non-parametric paired significance test, Wilcoxon signed-rank test, is utilized as we cannot assume the scores are normally distributed. We use two set $Q = \{q_1,q_2,\dotsc\}$ and $B = \{b_1,b_2,\dotsc\}$ to represent the ratings of ordinary questions and bad references, where $q_i$ and $b_i$ respectively represent the scores of $n$ rating criteria for an ordinary question and its related bad reference. For this experiment, we have 4 rating criteria as described in Table \ref{tbl:qg-rate-labels}.
We then compare the $p$-value produced by the significant test between $Q$ and $B$ with a selected threshold $\alpha$ to test whether the scores of ordinary questions are significantly higher than those of bad references. We apply the significance test on each worker, and the HITs from a worker with resulting $p < \alpha$ are kept. We choose $\alpha=0.05$ as our threshold as it is a common practice \cite{da,mrc-human-eval}. 

\subsubsection{Structure of HIT}
The questions to be evaluated as well as their passages and answers are generated on the HotpotQA test set by 11 various systems, including one system called ``Human'' that can simulate the human performance, and 10 neural-network-based QG systems, the details of which will be introduced in Section \ref{sec:qg-dataset-and-systems}. 

\begin{figure*}[ht]
    \centering
    \includegraphics[width=0.8\linewidth]{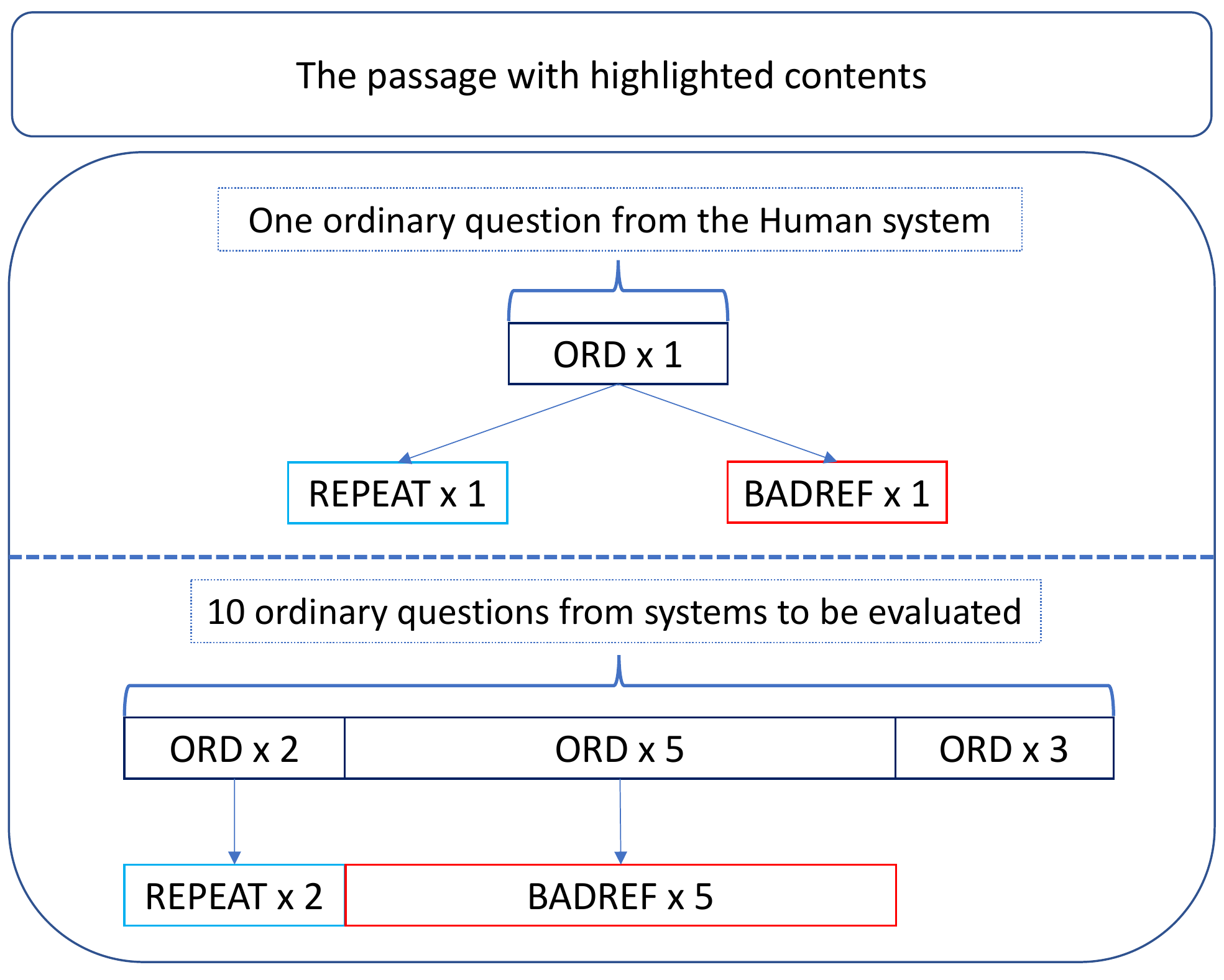}
    \caption{The structure of a single HIT in the QG evaluation experiment, where a HIT contains a certain passage, 11 system-generated questions and 9 variant questions for the purpose of controlling the quality. Meanwhile, ORD, REPEAT and BADREF respectively represent ordinary, repeat and bad reference questions.}
    \label{fig:qg-hit-structure}
\end{figure*}

For other tasks involving crowd-sourced human evaluation, a single HIT is made up of 100 items to rate \citep{mrc-human-eval}. However, HITs with similar size are inappropriate in this case as a passage containing several sentences should be provided for workers, and a 100-item HIT means a highly oversized workload for an individual. The reading quantity in a single HIT is one of concern as our preliminary experiment shows that a HIT with too many contents to read can significantly decrease the workers' efficiency. 
Instead, we organize the structure of HITs in the QG evaluation experiment as follows:
\begin{itemize}
    \item 1 original question, 1 repeat and 1 bad reference from the Human system (comprising a total of 3 questions);
    \item 2 original questions and their repeats from 2 of the 10 neural QG systems (comprising a total of 4 questions);
    \item 5 original questions and their bad references from the other 5 of the 10 normal systems (comprising a total of 10 questions);
    \item 3 original questions from the remaining 3 of the 10 normal systems (comprising a total of 3 questions).
\end{itemize}
where all these questions in one HIT share the identical passage and the correct answer. In other words, each HIT in the QG evaluation experiment consists of 20 items to rate, including: (a) 11 ordinary system-generated questions; (b) 6 bad reference questions corresponding to 6 of these 11; (c) 3 exact repeats corresponding to 3 of these 11. Figure \ref{fig:qg-hit-structure} provides the detailed structure of a HIT, where ORD=ordinary question, REPEAT=repeat question and BADREF=bad reference question. Although the hierarchical structure in Figure \ref{fig:qg-hit-structure} organizes the 20 items in a certain order, they will be fully shuffled before the deployment.

%% file: 4-Experiment-Results.tex
\subsection{Experiment results}
\label{chp-04-sec-04-experiment-result}
In this section, we design an experiment to investigate our proposed human evaluation method using the crowd-sourcing platform Amazon Mechanical Turk (AMT) (\url{www.mturk.com}), and we report the details of experiments, such as the pass rate of workers and the cost of deploying experiments. We also report the human score of QG systems at the system-level based on the collected data, and deploy a self-replication experiment to inquire into the reliability of this proposed human evaluation method.

\input{subsections/04-experiment-result}

%% file: subsections/04-experiment-result.tex
\input{subsections/04-sub-01-stats}

\input{subsections/04-sub-02-human-scores}

\input{subsections/04-sub-03-system-consistency}

%% file: subsections/04-sub-01-stats.tex
\subsubsection{Workers and HITs}
Two runs of experiments are deployed on the AMT platform, where the second run is designed to serve as a self-replication experiment to ensure the reliability of experimental findings. We then compute the correlation between the human scores of two runs at the system-level to examine the consistency of our method, which will be introduced in Section \ref{exp-result-subsec-system-consistency}. The HITs in the two experimental runs are randomly sampled from a HIT pool, which is generated as the outputs from the aforementioned QG systems. Table \ref{tbl:qg-passrate-and-worktime} provides statistical information with regard to the data of workers and HITs collected from our human evaluation experiments.

\begin{table*}[t]
    \centering
    \caption{Statistical information of the collected experiment data.}
    \input{tables/tbl-qg-passrate-and-worktime}
    \label{tbl:qg-passrate-and-worktime}
\end{table*}

Table \ref{tbl:qg-passrate} shows the numbers of human raters who participate in the QG evaluation experiment on the AMT platform, who passed the quality control and their pass rate for two distinct runs. The quality control method is as described in Section \ref{exp-design-subsec-qc}. The number of HITs before and after quality control, as well as the pass rate are also reported. For the first run, we collected $334$ passed HITs resulting in a total of $18,704$ valid ratings. Specifically, a non-human system on average received $1,603$ ratings and the human system received $2,672$ ratings, which is a sufficient sample size since it exceeds the minimum acceptable number (approximately $385$) according to the related research of statistical power in MT \cite{translationese}. 

Table \ref{tbl:qg-worktime} shows the average duration of a HIT and how many HITs a worker takes on the average according to the influence of the quality control method for both runs. Human raters whose HITs pass the quality control threshold usually spend a longer time completing a HIT than raters of failed HITs.


\subsubsection{Cost of the experiment}
Similar to other crowd-sourcing human experiments on the AMT platform \cite{da,mrc-human-eval}, a worker who passed the quality control was paid $0.99$ USD per completed HIT. This entire experiment cost less than $700$ USD in total. For research using our proposed evaluation method in the future, the total cost should be approximately half of this since we ran the experiment an additional time to investigate reliability, which generally is not required.
The resulting correlation between the system scores of the two separate data collection runs was $r=0.955$, sufficient to ensure reliability of results. Failed workers were often still paid for their time, where they could claim to have made an honest attempt at the HIT. Only obvious attempts to game the HITs are rejected.
In general, according to the cost of our first data collection run, assessing a QG system with nearly $1,600$ valid ratings in fact costed about $30$ USD (total cost $334$ USD $\div$ $11$ models $\approx$ $30.4$ USD). However, the experimental cost in future research may vary, depending on the sample size of collected data.


%% file: tables/tbl-qg-passrate-and-worktime.tex
\begin{subtable}[t]{0.9\linewidth}
    \centering
    \caption{The numbers of both workers and HITs before and after the quality-controlling mechanism as well as their pass rates for two runs.}
    \begin{tabularx}{\linewidth}{ccYYYcYYY}
    \hline
    \multirow{2}[4]{*}{Experiment} &       & \multicolumn{3}{c}{Worker} &       & \multicolumn{3}{c}{HIT} \bigstrut\\
    \cline{3-5}\cline{7-9}      &       & Passed & Total & Pass rate &       & Passed & Total & Pass rate \bigstrut\\
    \cline{1-1}\cline{3-5}\cline{7-9}Run1  &       & 123   & 356   & 34.55\% &       & 334   & 786   & 42.49\% \bigstrut[t]\\
    Run2  &       & 105   & 283   & 37.10\% &       & 282   & 598   & 47.16\% \bigstrut[b]\\
    \hline
    \end{tabularx}
   
    \label{tbl:qg-passrate}
\end{subtable}

\bigskip

\begin{subtable}[t]{0.9\linewidth}
    \centering
    \caption{The average elapsed time per HIT needed to be completed in minutes, and the average number of HITs that a worker is assigned.}

    \begin{tabularx}{\linewidth}{ccYYYcYYY}
    \hline
    \multirow{3}[4]{*}{Experiment} &       & \multicolumn{3}{c}{Elapsed time} &       & \multicolumn{3}{c}{Assigned HIT} \bigstrut[t]\\
          &       & \multicolumn{3}{c}{(per HIT in minutes)} &       & \multicolumn{3}{c}{(per worker)} \bigstrut[b]\\
    \cline{3-5}\cline{7-9}      &       & Passed & Failed & Total &       & Passed & Failed & Total \bigstrut\\
    \cline{1-1}\cline{3-5}\cline{7-9}Run1  &       & 33.24 & 26.93 & 29.61 &       & 2.72  & 1.94  & 2.21 \bigstrut[t]\\
    Run2  &       & 38.68 & 25.79 & 31.87 &       & 2.69  & 1.78  & 2.11 \bigstrut[b]\\
    \hline
    \end{tabularx}
    
    \label{tbl:qg-worktime}
\end{subtable}

%% file: subsections/04-sub-02-human-scores.tex
\subsubsection{Human scores}

Human raters may have different scoring strategies, for example, some strict raters tend to give a lower score to the same question compared with other raters. Therefore, we use the average standardized ($z$) scores instead of the original score, in order to iron out differences resulting from different strategies. Equation \ref{eq:zscore} is the computation of the average standardized scores for each evaluation criterion and the overall score of a QG system:

\begin{equation}
\label{eq:zscore}
\begin{aligned}
z_q^c  &= \frac{r_q^{w} - \mu_w}{\sigma_w} \\
z^c &= \frac{1}{\left|Q\right|}\sum_{q\in Q}{z_q^c} \\
z &= \frac{1}{\left|C\right|}\sum_{c\in C}{z^c} \\
\end{aligned}
\end{equation}
where the standardized score $z_q^c$ on the criterion $c$ of a system-generated question $q$ is computed by its raw score $r_q^c$ and the mean $\mu_w$ and the standard deviation $\sigma_w$ of its rater $w$, $z^c$ is the system-level standardized score on the criterion $c$ of a QG system, $Q$ is the set consisting of all rated questions ($q$) belonging to the QG system, and the overall average standardized scores $z$ is computed by averaging the $z^c$ of all criteria ($C$).

\begin{table*}[t]
    \centering
    \caption{Human evaluation standardized $z$ scores of overall and all rating criteria in the first run, where a bold value indicates the system receives the highest score among systems except the Human system, and $N$ indicates the number of evaluated questions of a system; systems (described in Section \ref{subsec:qg-systems}) are sorted by the overall score.}
    \input{tables/tbl-qg-zscores-r1}
    \label{tbl:qg-raw-and-z-r1}
\end{table*}

Table \ref{tbl:qg-raw-and-z-r1} shows the standardized human scores of all systems based on the ratings from all passed workers in the first run as well as the sample size $N$, where overall is the arithmetic mean of the scores of understandability, relevancy, answerability and appropriateness. A highlighted value indicates the system in the row outperforms every other system excluding the human Human question for that rating criterion.
For the calculation of standardized $z$ scores, the scores of bad references are not included, and for repeat questions the mean score of both evaluations for that question are combined into the final score.

As described in Table \ref{tbl:qg-raw-and-z-r1}, the Human system receives the best $z$ scores among all evaluation aspects, which is as expected since it consists of human-generated questions. 
For all QG systems excluding Human, BART$_{large}$ outperforms all other systems overall, and individually for understandability, relevancy and appropriateness. We also find that BART$_{base}$ somehow performs better than BART$_{large}$ at the answerability criterion. This is interesting as the performance of a model should generally increase if it is trained on a larger corpus. We think this implies that training models on a larger scale may potentially reduce the ability to generate high quality questions in terms of some aspects, namely answerability in this case. This is probably because a larger corpus may contain more noise which can negatively influence some aspects of a model, and it is worth investigating in future work.


%% file: tables/tbl-qg-zscores-r1.tex
\begin{tabularx}{\linewidth}{lccYYYY}
\hline
\multicolumn{1}{c}{System} & $N$ & \rotatebox{45}{Overall}   & \rotatebox{45}{Understandability} & \rotatebox{45}{Relevancy} & \rotatebox{45}{Answerability} & \rotatebox{45}{Appropriateness} \bigstrut\\
\hline
Human  & 668   & \phantom{$-$}0.322  & \phantom{$-$}0.164  & \phantom{$-$}0.262  & \phantom{$-$}0.435  & \phantom{$-$}0.429 \bigstrut[t] \\
BART$_{large}$ & 400   & \phantom{$-$}\textbf{0.308}  & \phantom{$-$}\textbf{0.155}  & \phantom{$-$}\textbf{0.255}  & \phantom{$-$}0.420  & \phantom{$-$}\textbf{0.403} \\
BART$_{base}$ & 401   & \phantom{$-$}0.290  & \phantom{$-$}0.135  & \phantom{$-$}0.234  & \phantom{$-$}\textbf{0.430}  & \phantom{$-$}0.360  \\
T5$_{base}$ & 395   & \phantom{$-$}0.226  & \phantom{$-$}0.051  & \phantom{$-$}0.241  & \phantom{$-$}0.395  & \phantom{$-$}0.217  \\
RNN   & 395   & \phantom{$-$}0.147  & $-$0.050  & \phantom{$-$}0.128  & \phantom{$-$}0.222  & \phantom{$-$}0.289  \\
Seq2Seq & 404   & \phantom{$-$}0.120  & $-$0.030  & \phantom{$-$}0.022  & \phantom{$-$}0.180  & \phantom{$-$}0.309  \\
T5$_{small}$ & 405   & \phantom{$-$}0.117  & $-$0.108  & \phantom{$-$}0.106  & \phantom{$-$}0.260  & \phantom{$-$}0.210  \\
Baseline$_{plus}$ & 408   & \phantom{$-$}0.076  & $-$0.133  & \phantom{$-$}0.076  & \phantom{$-$}0.196  & \phantom{$-$}0.165  \\
Seq2Seq$^*$ & 396   & \phantom{$-$}0.053  & $-$0.055  & $-$0.039  & \phantom{$-$}0.088  & \phantom{$-$}0.217  \\
Baseline & 396   & $-$0.008  & $-$0.186  & $-$0.032  & \phantom{$-$}0.155  & \phantom{$-$}0.032  \\
GPT-2 & 408   & $-$0.052  & $-$0.202  & $-$0.126  & \phantom{$-$}0.050  & \phantom{$-$}0.068  \bigstrut[b]\\
\hline
\end{tabularx}

%% file: subsections/04-sub-03-system-consistency.tex
\subsubsection{System consistency}
\label{exp-result-subsec-system-consistency}
To assess the reliability of the proposed human evaluation method, two distinct runs of the experiment are deployed with different human raters and HITs on the AMT platform. We think a robust evaluation method should be able to have a high correlation between the results of two independent experiment runs.

\begin{table*}[t]
    \centering
    \caption{Human evaluation standardized $z$ scores of overall and all rating criteria in the second run, where these systems follows the order in Table \ref{tbl:qg-raw-and-z-r1}, and $N$ indicates the number of evaluated questions of a system.}
\input{tables/tbl-qg-zscores-r2}
    \label{tbl:qg-raw-and-z-r2}
\end{table*}

\begin{table*}[ht]
    \centering
    \caption{The Pearson ($r$), Spearman ($\rho$) and Kendall's tau ($\tau$) correlations between the standardized $z$ scores of two runs of the experiment, including overall and four evaluation criteria.}
    \input{tables/tbl-qg-corr-r1r2}
    
    \label{tbl:qg-corr-r1r2}
\end{table*}

Table \ref{tbl:qg-raw-and-z-r2} shows the human evaluation results on the second run of our experiment, where the systems follows the order in the first run. We additionally compute the correlation coefficients between the standardized $z$ scores of both runs as shown in Table \ref{tbl:qg-corr-r1r2}, where $r$, $\rho$ and $\tau$ represent Pearson, Spearman and Kendall's tau correlation, respectively. We observe that the overall scores of two distinct experimental runs can reach $r=0.955$, while Person correlation of other evaluation criteria ranges from $0.865$ (Relevancy) to $0.957$ (Answerability). We believe such high correlation values are sufficient to indicate the robustness and reliability of this proposed human evaluation method.

%% file: tables/tbl-qg-zscores-r2.tex
\begin{tabularx}{\linewidth}{lccYYYY}
\hline
\multicolumn{1}{c}{System} & $N$ & \rotatebox{45}{Overall}   & \rotatebox{45}{Understandability} & \rotatebox{45}{Relevancy} & \rotatebox{45}{Answerability} & \rotatebox{45}{Appropriateness} \bigstrut\\
\hline

Human  & 564   & \phantom{$-$}0.316  & \phantom{$-$}0.188  & \phantom{$-$}0.279  & \phantom{$-$}0.386  & \phantom{$-$}0.410  \bigstrut[t]\\
BART$_{large}$ & 342   & \phantom{$-$}0.299  & \phantom{$-$}0.180  & \phantom{$-$}0.277  & \phantom{$-$}0.380  & \phantom{$-$}0.359  \\
BART$_{base}$ & 338   & \phantom{$-$}0.306  & \phantom{$-$}0.181  & \phantom{$-$}0.299  & \phantom{$-$}0.397  & \phantom{$-$}0.347  \\
T5$_{base}$ & 329   & \phantom{$-$}0.294  & \phantom{$-$}0.158  & \phantom{$-$}0.298  & \phantom{$-$}0.396  & \phantom{$-$}0.326  \\
RNN   & 342   & \phantom{$-$}0.060  & $-$0.040  & $-$0.008  & \phantom{$-$}0.072  & \phantom{$-$}0.217  \\
Seq2Seq & 332   & \phantom{$-$}0.086  & $-$0.053  & \phantom{$-$}0.064  & \phantom{$-$}0.115  & \phantom{$-$}0.217  \\
T5$_{small}$ & 340   & \phantom{$-$}0.157  & $-$0.012  & \phantom{$-$}0.166  & \phantom{$-$}0.248  & \phantom{$-$}0.224  \\
Baseline$_{plus}$ & 341   & \phantom{$-$}0.069  & $-$0.094  & \phantom{$-$}0.081  & \phantom{$-$}0.134  & \phantom{$-$}0.157  \\
Seq2Seq$^*$ & 348   & \phantom{$-$}0.083  & $-$0.014  & \phantom{$-$}0.077  & \phantom{$-$}0.104  & \phantom{$-$}0.163  \\
Baseline & 329   & $-$0.025  & $-$0.200  & $-$0.023  & \phantom{$-$}0.042  & \phantom{$-$}0.083  \\
GPT-2 & 343   & $-$0.047  & $-$0.122  & \phantom{$-$}0.000  & $-$0.036  & $-$0.031  \bigstrut[b]\\

\hline
\end{tabularx}

%% file: tables/tbl-qg-corr-r1r2.tex

\begin{tabularx}{0.9\linewidth}{ccYYYY}
\hline

& Overall  & Understandability & Relevancy & Answerability & Appropriateness \bigstrut\\
\hline

\multicolumn{1}{c}{$r$}  & 0.955  & 0.953  & 0.865  & 0.957  & 0.884  \bigstrut[t]\\
\multicolumn{1}{c}{$\rho$} & 0.882  & 0.891  & 0.718  & 0.882  & 0.845  \\
\multicolumn{1}{c}{$\tau$} & 0.745  & 0.709  & 0.527  & 0.745  & 0.709  \bigstrut[b]\\
\hline
\end{tabularx}

%% file: 5-Conclusion.tex
\section{Conclusion}

In this paper, we propose a new automatic evaluation metric -- QAScore, and a new crowd-sourcing human evaluation method for the task of question generation. Compared with other metrics, our metric can evaluate a question only according to its relevant passage and answer without the reference. 
QAScore utilizes the pretrained language model RoBERTa, and it evaluated a system-generated question by computing the cross entropy regarding the probability that RoBERTa can properly predict the masked words in the answer. 


We additionally propose a new crowd-sourced human evaluation method for the task of question generation. Each candidate question is evaluated on four various aspects: understandability, relevancy, answerability and appropriateness. To investigate the reliability of our method, we deployed a self-replication experiment whereby the correlation between the results from two independent runs is shown as high as $r=0.955$. We also provide a method of filtering out unreliable data from crowd-sourced workers. We introduce the structure of a HIT, the dataset we used and the involved QG systems to encourage the community to repeat our experiment.

According to the results of our human evaluation experiment, we further investigate how  performances of QAScore and other metrics. Results show that QAScore achieves the highest correlation with human judgements, which means QAScore can outperform existing QG evaluation metrics.



In conclusion, we propose an unsupervised reference-free automatic metric which correlates better with human judgements compared with other automatic metrics. In addition, we propose a crowd-sourced evaluation method for the question generation task which is highly robust and effective and it can be deployed within a limited budge of time and resources.

%% file: acl_latex.bbl
\begin{thebibliography}{56}
\expandafter\ifx\csname natexlab\endcsname\relax\def\natexlab#1{#1}\fi

\bibitem[{Banerjee and Lavie(2005)}]{metrics-meteor-1.0}
Satanjeev Banerjee and Alon Lavie. 2005.
\newblock \href {https://aclanthology.org/W05-0909} {{METEOR}: An automatic
  metric for {MT} evaluation with improved correlation with human judgments}.
\newblock In \emph{Proceedings of the {ACL} Workshop on Intrinsic and Extrinsic
  Evaluation Measures for Machine Translation and/or Summarization}, pages
  65--72, Ann Arbor, Michigan. Association for Computational Linguistics.

\bibitem[{Chen et~al.(2017)Chen, Fisch, Weston, and Bordes}]{chen2017reading}
Danqi Chen, Adam Fisch, Jason Weston, and Antoine Bordes. 2017.
\newblock Reading wikipedia to answer open-domain questions.
\newblock In \emph{Proceedings of the 55th Annual Meeting of the Association
  for Computational Linguistics (Volume 1: Long Papers)}, pages 1870--1879.

\bibitem[{Chen et~al.(2019)Chen, Wu, and Zaki}]{chen2020reinforcement}
Yu~Chen, Lingfei Wu, and Mohammed~J Zaki. 2019.
\newblock Reinforcement learning based graph-to-sequence model for natural
  question generation.
\newblock In \emph{International Conference on Learning Representations}.

\bibitem[{Chen et~al.(2020)Chen, Wu, and Zaki}]{qblue-use1}
Yu~Chen, Lingfei Wu, and Mohammed~J. Zaki. 2020.
\newblock Reinforcement learning based graph-to-sequence model for natural
  question generation.
\newblock In \emph{Proceedings of the 8th International Conference on Learning
  Representations}.

\bibitem[{Cho et~al.(2014)Cho, van Merri{\"e}nboer, Gulcehre, Bahdanau,
  Bougares, Schwenk, and Bengio}]{model-rnn}
Kyunghyun Cho, Bart van Merri{\"e}nboer, Caglar Gulcehre, Dzmitry Bahdanau,
  Fethi Bougares, Holger Schwenk, and Yoshua Bengio. 2014.
\newblock Learning phrase representations using rnn encoder--decoder for
  statistical machine translation.
\newblock In \emph{Proceedings of the 2014 Conference on Empirical Methods in
  Natural Language Processing (EMNLP)}, pages 1724--1734.

\bibitem[{Cho et~al.(2021)Cho, Zhang, Rao, Celikyilmaz, Xiong, Gao, Wang, and
  Dolan}]{qg-model-example3-ms-marco}
Woon~Sang Cho, Yizhe Zhang, Sudha Rao, Asli Celikyilmaz, Chenyan Xiong,
  Jianfeng Gao, Mengdi Wang, and Bill Dolan. 2021.
\newblock \href {https://aclanthology.org/2021.eacl-main.2} {Contrastive
  multi-document question generation}.
\newblock In \emph{Proceedings of the 16th Conference of the European Chapter
  of the Association for Computational Linguistics: Mainv Volume}, pages
  12--30, Online. Association for Computational Linguistics.

\bibitem[{Devlin et~al.(2019)Devlin, Chang, Lee, and Toutanova}]{model-bert}
Jacob Devlin, Ming-Wei Chang, Kenton Lee, and Kristina Toutanova. 2019.
\newblock \href {https://doi.org/10.18653/v1/N19-1423} {{BERT}: Pre-training of
  deep bidirectional transformers for language understanding}.
\newblock In \emph{Proceedings of the 2019 Conference of the North {A}merican
  Chapter of the Association for Computational Linguistics: Human Language
  Technologies, Volume 1 (Long and Short Papers)}, pages 4171--4186,
  Minneapolis, Minnesota. Association for Computational Linguistics.

\bibitem[{Du et~al.(2017)Du, Shao, and Cardie}]{du-etal-2017-learning}
Xinya Du, Junru Shao, and Claire Cardie. 2017.
\newblock \href {https://doi.org/10.18653/v1/P17-1123} {Learning to ask: Neural
  question generation for reading comprehension}.
\newblock In \emph{Proceedings of the 55th Annual Meeting of the Association
  for Computational Linguistics (Volume 1: Long Papers)}, pages 1342--1352,
  Vancouver, Canada. Association for Computational Linguistics.

\bibitem[{Graham(2015)}]{graham2015re}
Yvette Graham. 2015.
\newblock \href {https://doi.org/10.18653/v1/D15-1013} {Re-evaluating automatic
  summarization with {BLEU} and 192 shades of {ROUGE}}.
\newblock In \emph{Proceedings of the 2015 Conference on Empirical Methods in
  Natural Language Processing}, pages 128--137, Lisbon, Portugal. Association
  for Computational Linguistics.

\bibitem[{Graham et~al.(2016)Graham, Baldwin, Moffat, and Zobel}]{da}
Yvette Graham, Timothy Baldwin, Alistair Moffat, and Justin Zobel. 2016.
\newblock \href {https://doi.org/10.1017/S1351324915000339} {Can machine
  translation systems be evaluated by the crowd alone}.
\newblock \emph{Natural Language Engineering}, FirstView:1--28.

\bibitem[{Graham et~al.(2020)Graham, Haddow, and Koehn}]{translationese}
Yvette Graham, Barry Haddow, and Philipp Koehn. 2020.
\newblock \href {https://doi.org/10.18653/v1/2020.emnlp-main.6} {Statistical
  power and translationese in machine translation evaluation}.
\newblock In \emph{Proceedings of the 2020 Conference on Empirical Methods in
  Natural Language Processing (EMNLP)}, pages 72--81, Online. Association for
  Computational Linguistics.

\bibitem[{Graham and Liu(2016)}]{graham2016achieving}
Yvette Graham and Qun Liu. 2016.
\newblock Achieving accurate conclusions in evaluation of automatic machine
  translation metrics.
\newblock In \emph{Proceedings of the 2016 Conference of the North American
  Chapter of the Association for Computational Linguistics: Human Language
  Technologies}, pages 1--10.

\bibitem[{Ji et~al.(2022)Ji, Graham, Jones, Lyu, and Liu}]{ji2022achieving}
Tianbo Ji, Yvette Graham, Gareth Jones, Chenyang Lyu, and Qun Liu. 2022.
\newblock \href {https://doi.org/10.18653/v1/2022.acl-long.445} {Achieving
  reliable human assessment of open-domain dialogue systems}.
\newblock In \emph{Proceedings of the 60th Annual Meeting of the Association
  for Computational Linguistics (Volume 1: Long Papers)}, pages 6416--6437,
  Dublin, Ireland. Association for Computational Linguistics.

\bibitem[{Ji et~al.(2020)Ji, Graham, and Jones}]{mrc-human-eval}
Tianbo Ji, Yvette Graham, and Gareth~J.F. Jones. 2020.
\newblock \href {https://doi.org/10.1145/3343413.3377996} {\emph{Contrasting
  Human Opinion of Non-Factoid Question Answering with Automatic Evaluation}},
  page 348–352. Association for Computing Machinery, New York, NY, USA.

\bibitem[{Ji et~al.(2021)Ji, Lyu, Cao, and
  Cheng}]{model-hierarchical-seq2seq-and-context-switch}
Tianbo Ji, Chenyang Lyu, Zhichao Cao, and Peng Cheng. 2021.
\newblock \href {https://doi.org/10.3390/e23111449} {Multi-hop question
  generation using hierarchical encoding-decoding and context switch
  mechanism}.
\newblock \emph{Entropy}, 23(11).

\bibitem[{Jia et~al.(2021)Jia, Zhou, Sun, and Wu}]{qg-model-n-point-example1}
Xin Jia, Wenjie Zhou, Xu~Sun, and Yunfang Wu. 2021.
\newblock Eqg-race: Examination-type question generation.
\newblock In \emph{AAAI}.

\bibitem[{Kim et~al.(2019)Kim, Lee, Shin, and Jung}]{qg-model-example1-squad}
Yanghoon Kim, Hwanhee Lee, Joongbo Shin, and Kyomin Jung. 2019.
\newblock \href {https://doi.org/10.1609/aaai.v33i01.33016602} {Improving
  neural question generation using answer separation}.
\newblock \emph{Proceedings of the AAAI Conference on Artificial Intelligence},
  33(01):6602--6609.

\bibitem[{Ko{\v{c}}isk{\'y} et~al.(2018)Ko{\v{c}}isk{\'y}, Schwarz, Blunsom,
  Dyer, Hermann, Melis, and Grefenstette}]{free-answering-narrativeqa}
Tom{\'a}{\v{s}} Ko{\v{c}}isk{\'y}, Jonathan Schwarz, Phil Blunsom, Chris Dyer,
  Karl~Moritz Hermann, G{\'a}bor Melis, and Edward Grefenstette. 2018.
\newblock \href {https://doi.org/10.1162/tacl_a_00023} {The {N}arrative{QA}
  reading comprehension challenge}.
\newblock \emph{Transactions of the Association for Computational Linguistics},
  6:317--328.

\bibitem[{Lewis et~al.(2020{\natexlab{a}})Lewis, Liu, Goyal, Ghazvininejad,
  Mohamed, Levy, Stoyanov, and Zettlemoyer}]{model-bart}
Mike Lewis, Yinhan Liu, Naman Goyal, Marjan Ghazvininejad, Abdelrahman Mohamed,
  Omer Levy, Veselin Stoyanov, and Luke Zettlemoyer. 2020{\natexlab{a}}.
\newblock \href {https://doi.org/10.18653/v1/2020.acl-main.703} {{BART}:
  Denoising sequence-to-sequence pre-training for natural language generation,
  translation, and comprehension}.
\newblock In \emph{Proceedings of the 58th Annual Meeting of the Association
  for Computational Linguistics}, pages 7871--7880, Online. Association for
  Computational Linguistics.

\bibitem[{Lewis et~al.(2020{\natexlab{b}})Lewis, Perez, Piktus, Petroni,
  Karpukhin, Goyal, K\"{u}ttler, Lewis, Yih, Rockt\"{a}schel, Riedel, and
  Kiela}]{qblue-use2}
Patrick Lewis, Ethan Perez, Aleksandra Piktus, Fabio Petroni, Vladimir
  Karpukhin, Naman Goyal, Heinrich K\"{u}ttler, Mike Lewis, Wen-tau Yih, Tim
  Rockt\"{a}schel, Sebastian Riedel, and Douwe Kiela. 2020{\natexlab{b}}.
\newblock \href
  {https://proceedings.neurips.cc/paper/2020/file/6b493230205f780e1bc26945df7481e5-Paper.pdf}
  {Retrieval-augmented generation for knowledge-intensive nlp tasks}.
\newblock In \emph{Advances in Neural Information Processing Systems},
  volume~33, pages 9459--9474. Curran Associates, Inc.

\bibitem[{Lewis et~al.(2021)Lewis, Wu, Liu, Minervini, Küttler, Piktus,
  Stenetorp, and Riedel}]{lewis2021paq}
Patrick Lewis, Yuxiang Wu, Linqing Liu, Pasquale Minervini, Heinrich Küttler,
  Aleksandra Piktus, Pontus Stenetorp, and Sebastian Riedel. 2021.
\newblock \href {http://arxiv.org/abs/2102.07033} {Paq: 65 million
  probably-asked questions and what you can do with them}.

\bibitem[{Li et~al.(2021)Li, Qu, Yan, Zhou, and
  Cheng}]{10.1007/978-3-030-82136-4_13}
Jianbin Li, Ketong Qu, Jingchen Yan, Liting Zhou, and Long Cheng. 2021.
\newblock Tebc-net: An effective relation extraction approach for simple
  question answering over knowledge graphs.
\newblock In \emph{Knowledge Science, Engineering and Management}, pages
  154--165, Cham. Springer International Publishing.

\bibitem[{Lin(2004)}]{metrics-rouge}
Chin-Yew Lin. 2004.
\newblock \href {https://www.aclweb.org/anthology/W04-1013} {{ROUGE}: A package
  for automatic evaluation of summaries}.
\newblock In \emph{Text Summarization Branches Out}, pages 74--81, Barcelona,
  Spain. Association for Computational Linguistics.

\bibitem[{Liu et~al.(2020)Liu, Wei, Niu, Chen, and
  He}]{qg-model-n-point-example3}
Bang Liu, Haojie Wei, Di~Niu, Haolan Chen, and Yancheng He. 2020.
\newblock \href {https://doi.org/10.1145/3366423.3380270} {Asking questions the
  human way: Scalable question-answer generation from text corpus}.
\newblock In \emph{Proceedings of The Web Conference 2020}, WWW '20, page
  2032–2043, New York, NY, USA. Association for Computing Machinery.

\bibitem[{Liu et~al.(2019)Liu, Ott, Goyal, Du, Joshi, Chen, Levy, Lewis,
  Zettlemoyer, and Stoyanov}]{model-roberta}
Yinhan Liu, Myle Ott, Naman Goyal, Jingfei Du, Mandar Joshi, Danqi Chen, Omer
  Levy, Mike Lewis, Luke Zettlemoyer, and Veselin Stoyanov. 2019.
\newblock \href {http://arxiv.org/abs/1907.11692} {Roberta: {A} robustly
  optimized {BERT} pretraining approach}.
\newblock \emph{CoRR}, abs/1907.11692.

\bibitem[{Lyu et~al.(2022)Lyu, Foster, and Graham}]{lyu-etal-2022-extending}
Chenyang Lyu, Jennifer Foster, and Yvette Graham. 2022.
\newblock \href {https://doi.org/10.18653/v1/2022.insights-1.4} {Extending the
  scope of out-of-domain: Examining {QA} models in multiple subdomains}.
\newblock In \emph{Proceedings of the Third Workshop on Insights from Negative
  Results in NLP}, pages 24--37, Dublin, Ireland. Association for Computational
  Linguistics.

\bibitem[{Lyu et~al.(2021)Lyu, Shang, Graham, Foster, Jiang, and
  Liu}]{lyu2021improving}
Chenyang Lyu, Lifeng Shang, Yvette Graham, Jennifer Foster, Xin Jiang, and Qun
  Liu. 2021.
\newblock \href {https://doi.org/10.18653/v1/2021.emnlp-main.340} {Improving
  unsupervised question answering via summarization-informed question
  generation}.
\newblock In \emph{Proceedings of the 2021 Conference on Empirical Methods in
  Natural Language Processing}, pages 4134--4148, Online and Punta Cana,
  Dominican Republic. Association for Computational Linguistics.

\bibitem[{Ma et~al.(2020)Ma, Zhu, Zhou, and Li}]{qg-model-n-point-example4}
Xiyao Ma, Qile Zhu, Yanlin Zhou, and Xiaolin Li. 2020.
\newblock \href {https://doi.org/10.1609/aaai.v34i05.6366} {Improving question
  generation with sentence-level semantic matching and answer position
  inferring}.
\newblock \emph{Proceedings of the AAAI Conference on Artificial Intelligence},
  34(05):8464--8471.

\bibitem[{Narayan et~al.(2020)Narayan, Sim{\~{o}}es, Ma, Craighead, and
  McDonald}]{qg-model-n-point-example5}
Shashi Narayan, Gon{\c{c}}alo Sim{\~{o}}es, Ji~Ma, Hannah Craighead, and
  Ryan~T. McDonald. 2020.
\newblock \href {http://arxiv.org/abs/2004.11026} {{QURIOUS:} question
  generation pretraining for text generation}.
\newblock \emph{CoRR}, abs/2004.11026.

\bibitem[{Nema and Khapra(2018)}]{metrics-answerability}
Preksha Nema and Mitesh~M. Khapra. 2018.
\newblock \href {https://doi.org/10.18653/v1/D18-1429} {Towards a better metric
  for evaluating question generation systems}.
\newblock In \emph{Proceedings of the 2018 Conference on Empirical Methods in
  Natural Language Processing}, pages 3950--3959, Brussels, Belgium.
  Association for Computational Linguistics.

\bibitem[{Pan et~al.(2019)Pan, Lei, Chua, and Kan}]{qg_survey_liangmingpan}
Liangming Pan, Wenqiang Lei, Tat{-}Seng Chua, and Min{-}Yen Kan. 2019.
\newblock \href {http://arxiv.org/abs/1905.08949} {Recent advances in neural
  question generation}.
\newblock \emph{CoRR}, abs/1905.08949.

\bibitem[{Pan et~al.(2020{\natexlab{a}})Pan, Xie, Feng, Chua, and
  Kan}]{pan-etal-2020-semantic}
Liangming Pan, Yuxi Xie, Yansong Feng, Tat-Seng Chua, and Min-Yen Kan.
  2020{\natexlab{a}}.
\newblock \href {https://doi.org/10.18653/v1/2020.acl-main.135} {Semantic
  graphs for generating deep questions}.
\newblock In \emph{Proceedings of the 58th Annual Meeting of the Association
  for Computational Linguistics}, pages 1463--1475, Online. Association for
  Computational Linguistics.

\bibitem[{Pan et~al.(2020{\natexlab{b}})Pan, Xie, Feng, Chua, and
  Kan}]{model-Att-GGNN}
Liangming Pan, Yuxi Xie, Yansong Feng, Tat-Seng Chua, and Min-Yen Kan.
  2020{\natexlab{b}}.
\newblock \href {https://doi.org/10.18653/v1/2020.acl-main.135} {Semantic
  graphs for generating deep questions}.
\newblock In \emph{Proceedings of the 58th Annual Meeting of the Association
  for Computational Linguistics}, pages 1463--1475, Online. Association for
  Computational Linguistics.

\bibitem[{Papineni et~al.(2002{\natexlab{a}})Papineni, Roukos, Ward, and
  Zhu}]{papineni-etal-2002-bleu}
Kishore Papineni, Salim Roukos, Todd Ward, and Wei-Jing Zhu.
  2002{\natexlab{a}}.
\newblock \href {https://doi.org/10.3115/1073083.1073135} {{B}leu: a method for
  automatic evaluation of machine translation}.
\newblock In \emph{Proceedings of the 40th Annual Meeting of the Association
  for Computational Linguistics}, pages 311--318, Philadelphia, Pennsylvania,
  USA. Association for Computational Linguistics.

\bibitem[{Papineni et~al.(2002{\natexlab{b}})Papineni, Roukos, Ward, and
  Zhu}]{metrics-bleu}
Kishore Papineni, Salim Roukos, Todd Ward, and Wei-Jing Zhu.
  2002{\natexlab{b}}.
\newblock \href {https://doi.org/10.3115/1073083.1073135} {Bleu: A method for
  automatic evaluation of machine translation}.
\newblock In \emph{Proceedings of the 40th Annual Meeting on Association for
  Computational Linguistics}, ACL '02, pages 311--318, Stroudsburg, PA, USA.
  Association for Computational Linguistics.

\bibitem[{Puri et~al.(2020)Puri, Spring, Shoeybi, Patwary, and
  Catanzaro}]{puri-etal-2020-training-synthetic}
Raul Puri, Ryan Spring, Mohammad Shoeybi, Mostofa Patwary, and Bryan Catanzaro.
  2020.
\newblock \href {https://doi.org/10.18653/v1/2020.emnlp-main.468} {Training
  question answering models from synthetic data}.
\newblock In \emph{Proceedings of the 2020 Conference on Empirical Methods in
  Natural Language Processing (EMNLP)}, pages 5811--5826, Online. Association
  for Computational Linguistics.

\bibitem[{Radford et~al.(2019)Radford, Wu, Child, Luan, Amodei, and
  Sutskever}]{model-gpt-2}
Alec Radford, Jeff Wu, Rewon Child, David Luan, Dario Amodei, and Ilya
  Sutskever. 2019.
\newblock Language models are unsupervised multitask learners.

\bibitem[{Raffel et~al.(2020)Raffel, Shazeer, Roberts, Lee, Narang, Matena,
  Zhou, Li, and Liu}]{model-t5}
Colin Raffel, Noam Shazeer, Adam Roberts, Katherine Lee, Sharan Narang, Michael
  Matena, Yanqi Zhou, Wei Li, and Peter~J. Liu. 2020.
\newblock \href {http://jmlr.org/papers/v21/20-074.html} {Exploring the limits
  of transfer learning with a unified text-to-text transformer}.
\newblock \emph{Journal of Machine Learning Research}, 21(140):1--67.

\bibitem[{Rajpurkar et~al.(2016)Rajpurkar, Zhang, Lopyrev, and
  Liang}]{rajpurkar-etal-2016-squad1.1}
Pranav Rajpurkar, Jian Zhang, Konstantin Lopyrev, and Percy Liang. 2016.
\newblock \href {https://doi.org/10.18653/v1/D16-1264} {{SQ}u{AD}: 100,000+
  questions for machine comprehension of text}.
\newblock In \emph{Proceedings of the 2016 Conference on Empirical Methods in
  Natural Language Processing}, pages 2383--2392, Austin, Texas. Association
  for Computational Linguistics.

\bibitem[{Reiter(2018)}]{review_BLEU}
Ehud Reiter. 2018.
\newblock \href {https://doi.org/10.1162/coli_a_00322} {{A Structured Review of
  the Validity of BLEU}}.
\newblock \emph{Computational Linguistics}, 44(3):393--401.

\bibitem[{Ren and Zhu(2020)}]{qg-model-n-point-example2}
Siyu Ren and Kenny~Q. Zhu. 2020.
\newblock \href {http://arxiv.org/abs/2004.09853} {Knowledge-driven distractor
  generation for cloze-style multiple choice questions}.
\newblock \emph{CoRR}, abs/2004.09853.

\bibitem[{Saha et~al.(2018)Saha, Aralikatte, Khapra, and
  Sankaranarayanan}]{span-extraction-duorc}
Amrita Saha, Rahul Aralikatte, Mitesh~M. Khapra, and Karthik Sankaranarayanan.
  2018.
\newblock \href {https://doi.org/10.18653/v1/P18-1156} {{D}uo{RC}: Towards
  complex language understanding with paraphrased reading comprehension}.
\newblock In \emph{Proceedings of the 56th Annual Meeting of the Association
  for Computational Linguistics (Volume 1: Long Papers)}, pages 1683--1693,
  Melbourne, Australia. Association for Computational Linguistics.

\bibitem[{Sellam et~al.(2020)Sellam, Das, and Parikh}]{metric-bleurt}
Thibault Sellam, Dipanjan Das, and Ankur Parikh. 2020.
\newblock \href {https://doi.org/10.18653/v1/2020.acl-main.704} {{BLEURT}:
  Learning robust metrics for text generation}.
\newblock In \emph{Proceedings of the 58th Annual Meeting of the Association
  for Computational Linguistics}, pages 7881--7892, Online. Association for
  Computational Linguistics.

\bibitem[{Shin et~al.(2020)Shin, Razeghi, Logan~IV, Wallace, and
  Singh}]{roberta-ability}
Taylor Shin, Yasaman Razeghi, Robert~L. Logan~IV, Eric Wallace, and Sameer
  Singh. 2020.
\newblock \href {https://doi.org/10.18653/v1/2020.emnlp-main.346}
  {{A}uto{P}rompt: {E}liciting {K}nowledge from {L}anguage {M}odels with
  {A}utomatically {G}enerated {P}rompts}.
\newblock In \emph{Proceedings of the 2020 Conference on Empirical Methods in
  Natural Language Processing (EMNLP)}, pages 4222--4235, Online. Association
  for Computational Linguistics.

\bibitem[{Sutskever et~al.(2014)Sutskever, Vinyals, and Le}]{NIPS2014_seq2seq}
Ilya Sutskever, Oriol Vinyals, and Quoc~V Le. 2014.
\newblock \href
  {https://proceedings.neurips.cc/paper/2014/file/a14ac55a4f27472c5d894ec1c3c743d2-Paper.pdf}
  {Sequence to sequence learning with neural networks}.
\newblock In \emph{Advances in Neural Information Processing Systems},
  volume~27. Curran Associates, Inc.

\bibitem[{Trischler et~al.(2017)Trischler, Wang, Yuan, Harris, Sordoni,
  Bachman, and Suleman}]{trischler-etal-2017-newsqa}
Adam Trischler, Tong Wang, Xingdi Yuan, Justin Harris, Alessandro Sordoni,
  Philip Bachman, and Kaheer Suleman. 2017.
\newblock \href {https://doi.org/10.18653/v1/W17-2623} {{N}ews{QA}: A machine
  comprehension dataset}.
\newblock In \emph{Proceedings of the 2nd Workshop on Representation Learning
  for {NLP}}, pages 191--200, Vancouver, Canada. Association for Computational
  Linguistics.

\bibitem[{Wang et~al.(2020)Wang, Xu, Lin, Zheng, and
  Shen}]{qg-model-example2-hotpotqa}
Liuyin Wang, Zihan Xu, Zibo Lin, Haitao Zheng, and Ying Shen. 2020.
\newblock \href {https://doi.org/10.18653/v1/2020.coling-main.452}
  {Answer-driven deep question generation based on reinforcement learning}.
\newblock In \emph{Proceedings of the 28th International Conference on
  Computational Linguistics}, pages 5159--5170, Barcelona, Spain (Online).
  International Committee on Computational Linguistics.

\bibitem[{Wu et~al.(2016)Wu, Schuster, Chen, Le, Norouzi, Macherey, Krikun,
  Cao, Gao, Macherey, Klingner, Shah, Johnson, Liu, Kaiser, Gouws, Kato, Kudo,
  Kazawa, Stevens, Kurian, Patil, Wang, Young, Smith, Riesa, Rudnick, Vinyals,
  Corrado, Hughes, and Dean}]{metrics-gleu}
Yonghui Wu, Mike Schuster, Zhifeng Chen, Quoc~V. Le, Mohammad Norouzi, Wolfgang
  Macherey, Maxim Krikun, Yuan Cao, Qin Gao, Klaus Macherey, Jeff Klingner,
  Apurva Shah, Melvin Johnson, Xiaobing Liu, Lukasz Kaiser, Stephan Gouws,
  Yoshikiyo Kato, Taku Kudo, Hideto Kazawa, Keith Stevens, George Kurian,
  Nishant Patil, Wei Wang, Cliff Young, Jason Smith, Jason Riesa, Alex Rudnick,
  Oriol Vinyals, Greg Corrado, Macduff Hughes, and Jeffrey Dean. 2016.
\newblock \href {http://arxiv.org/abs/1609.08144} {Google's neural machine
  translation system: Bridging the gap between human and machine translation}.
\newblock \emph{CoRR}, abs/1609.08144.

\bibitem[{Xie et~al.(2020)Xie, Pan, Wang, Kan, and
  Feng}]{xie-etal-2020-exploring}
Yuxi Xie, Liangming Pan, Dongzhe Wang, Min-Yen Kan, and Yansong Feng. 2020.
\newblock \href {https://doi.org/10.18653/v1/2020.coling-main.228} {Exploring
  question-specific rewards for generating deep questions}.
\newblock In \emph{Proceedings of the 28th International Conference on
  Computational Linguistics}, pages 2534--2546, Barcelona, Spain (Online).
  International Committee on Computational Linguistics.

\bibitem[{Xu et~al.(2022)Xu, Wang, Yu, Ritchie, Yao, Wu, Zhang, Li, Bradford,
  Sun, Hoang, Sang, Hou, Ma, Yang, Peng, Yu, and
  Warschauer}]{xu-etal-2022-fantastic}
Ying Xu, Dakuo Wang, Mo~Yu, Daniel Ritchie, Bingsheng Yao, Tongshuang Wu, Zheng
  Zhang, Toby Li, Nora Bradford, Branda Sun, Tran Hoang, Yisi Sang, Yufang Hou,
  Xiaojuan Ma, Diyi Yang, Nanyun Peng, Zhou Yu, and Mark Warschauer. 2022.
\newblock \href {https://doi.org/10.18653/v1/2022.acl-long.34} {Fantastic
  questions and where to find them: {F}airytale{QA} {--} an authentic dataset
  for narrative comprehension}.
\newblock In \emph{Proceedings of the 60th Annual Meeting of the Association
  for Computational Linguistics (Volume 1: Long Papers)}, pages 447--460,
  Dublin, Ireland. Association for Computational Linguistics.

\bibitem[{Yang et~al.(2018)Yang, Qi, Zhang, Bengio, Cohen, Salakhutdinov, and
  Manning}]{qg-hotpotqa}
Zhilin Yang, Peng Qi, Saizheng Zhang, Yoshua Bengio, William~W. Cohen, Ruslan
  Salakhutdinov, and Christopher~D. Manning. 2018.
\newblock {HotpotQA}: A dataset for diverse, explainable multi-hop question
  answering.
\newblock In \emph{Conference on Empirical Methods in Natural Language
  Processing ({EMNLP})}.

\bibitem[{Yuan et~al.(2017)Yuan, Wang, Gulcehre, Sordoni, Bachman, Zhang,
  Subramanian, and Trischler}]{qg-bleu-doubt}
Xingdi Yuan, Tong Wang, Caglar Gulcehre, Alessandro Sordoni, Philip Bachman,
  Saizheng Zhang, Sandeep Subramanian, and Adam Trischler. 2017.
\newblock \href {https://doi.org/10.18653/v1/W17-2603} {Machine comprehension
  by text-to-text neural question generation}.
\newblock In \emph{Proceedings of the 2nd Workshop on Representation Learning
  for {NLP}}, pages 15--25, Vancouver, Canada. Association for Computational
  Linguistics.

\bibitem[{Zhang et~al.(2020{\natexlab{a}})Zhang, Kishore, Wu, Weinberger, and
  Artzi}]{metric-bert-score}
Tianyi Zhang, Varsha Kishore, Felix Wu, Kilian~Q. Weinberger, and Yoav Artzi.
  2020{\natexlab{a}}.
\newblock \href {https://openreview.net/forum?id=SkeHuCVFDr} {Bertscore:
  Evaluating text generation with bert}.
\newblock In \emph{International Conference on Learning Representations}.

\bibitem[{Zhang et~al.(2020{\natexlab{b}})Zhang, Zhao, and
  Wang}]{zhang2020machine}
Zhuosheng Zhang, Hai Zhao, and Rui Wang. 2020{\natexlab{b}}.
\newblock \href {http://arxiv.org/abs/2005.06249} {Machine reading
  comprehension: The role of contextualized language models and beyond}.

\bibitem[{Zhou and Zhang(2021)}]{qg-data-aug-for-qa1}
Shuohua Zhou and Yanping Zhang. 2021.
\newblock \href {https://doi.org/10.3390/app112311251} {Datlmedqa: A data
  augmentation and transfer learning based solution for medical question
  answering}.
\newblock \emph{Applied Sciences}, 11(23).

\bibitem[{Zhu et~al.(2021)Zhu, Lei, Wang, Zheng, Poria, and
  Chua}]{zhu2021retrieving}
Fengbin Zhu, Wenqiang Lei, Chao Wang, Jianming Zheng, Soujanya Poria, and
  Tat-Seng Chua. 2021.
\newblock Retrieving and reading: A comprehensive survey on open-domain
  question answering.
\newblock \emph{arXiv preprint arXiv:2101.00774}.

\end{thebibliography}
